%% file: main_camera_ready.tex
\documentclass[11pt]{article}

\usepackage[preprint]{acl}

\usepackage{times}
\usepackage{latexsym}

\usepackage[T1]{fontenc}

\usepackage[utf8]{inputenc}

\usepackage{microtype}

\usepackage{inconsolata}

\usepackage{graphicx}

\usepackage[noend]{algpseudocode} 
\usepackage[ruled, vlined]{algorithm2e}
\usepackage{amsmath,amssymb,amsfonts}
\usepackage{booktabs}
\usepackage{enumitem}
\usepackage{xcolor}
\usepackage{caption}
\usepackage{subcaption}
\usepackage{wrapfig}
\usepackage{mathtools}
\usepackage[most]{tcolorbox}
\usepackage[edges]{forest}
\usepackage{adjustbox}
\usepackage{tikz}
\usetikzlibrary{trees,positioning,shapes,shadows,arrows.meta}

\definecolor{cat}{HTML}{0B6FA4}
\definecolor{sub}{HTML}{3FA2DA}

\forestset{
tax/.style={
for tree={
draw, rounded corners, align=center, inner sep=3pt, outer sep=2pt,
edge={-latex},
s sep=7pt, l sep=8pt,
font=\small
}
},
cat/.style={fill=cat!15, draw=cat},
subcat/.style={fill=sub!15, draw=sub}
}
\usepackage{xcolor}
\usepackage{soul}
\sethlcolor{pink}
\usepackage{multirow}
\usepackage[most]{tcolorbox}  
\usepackage{capt-of}          

\title{JTPRO: A Joint Tool–Prompt Reflective Optimization Framework for Language Agents}

\author{
  \textbf{Sandip Ghoshal},
  \textbf{Anshul Mittal},
  \textbf{Jyotika Singh},
  \textbf{Miguel Ballesteros},
\\    \textbf{Weiyi Sun},
 \textbf{Fang Tu},
 \textbf{Shailender Singh},
  \textbf{Yassine Benajiba},
\\  \textbf{Sujeeth Bharadwaj},
  \textbf{Fahad Shah},
  \textbf{Sujith Ravi},
  \textbf{Dan Roth}
\\
 Oracle AI
\\
  \small{
    \textbf{Correspondence:} \href{mailto:sandip.ghoshal@oracle.com}{sandip.ghoshal@oracle.com}
  }
}

\begin{document}
\maketitle

\begin{abstract}
Large language model (LLM) agents augmented with external tools often struggle as number of tools grow large and become domain-specific. In such settings, ambiguous tool descriptions and under-specified agent instructions frequently lead to tool mis-selection and incorrect slot/value instantiation. We hypothesize that this is due to two root causes: generic, one-size-fits-all prompts that ignore tool-specific nuances, and underspecified tool schemas that lack clear guidance on when and how to use each tool and how to format its parameters. We introduce \textbf{J}oint \textbf{T}ool-\textbf{P}rompt \textbf{R}eflective \textbf{O}ptimization (\textbf{JTPRO}), a framework for improving tool-calling reliability in \emph{trace-supervised} settings by iteratively using rollout-driven reflection to co-optimize global instructions and per-tool schema/argument descriptions for accurate tool selection and argument instantiation in large tool inventories. JTPRO is designed to preserve only tool-local cues needed for correct disambiguation and slot filling. We evaluate JTPRO across multi-tool benchmarks, which account for different number of tools using three metrics: Tool Selection Accuracy (TSA), Slot Filling Accuracy(SFA), and Overall Success Rate(OSR) (correct tool + correct slots + correct values). JTPRO consistently outperforms strong baselines, including CoT-style agents, and reflective prompt optimizers such as GEPA by 5\%–20\% (relative) on OSR. Ablations show that joint optimization of instructions and tool schemas is more effective and robust than optimizing either component in isolation.
\end{abstract}

\input{sections/camera_ready/1-introduction}

\input{sections/camera_ready/2-relatedwork}
\input{sections/camera_ready/3-probstatement}

\input{sections/camera_ready/4-technique}

\input{sections/camera_ready/5-evalsetup}

\input{sections/camera_ready/6-resultsanalysis}
\input{sections/camera_ready/7-conclusion}
\bibliography{custom}
\newpage
\appendix
\input{sections/camera_ready/9_appendix}

\end{document}

%% file: sections/camera_ready/1-introduction.tex
\input{figures_camera_ready/tex/fig1_jtpro_gains}
\input{figures_camera_ready/tex/fig2_rag_norag}


\section{Introduction}
Tool-augmented large language model (LLM) \cite{Vaswani2017attention} agents extend their capabilities by invoking external tools for specialized operations and up-to-date information \cite{survey_wang_2024} and are an important real-world application \cite{S2023} across domains \cite{agenticaidomain, meghwani2025hard, s-etal-2025-llms}. In this work, we focus specifically on \emph{trace-supervised tool-calling settings}, where the objective is reliable call-level execution: (i) select the correct tool among many conflicting options, (ii) instantiate correct arguments from natural language requests; both suffer when tool/slot descriptions are ambiguous or underspecified \cite{qin2023toolllmfacilitatinglargelanguage}. \textbf{\autoref{fig:tool_scaling_and_slot_filling}} quantifies this scaling failure on ToolACE~\cite{liu2025toolace}: (a) tool selection accuracy drops as the tool universe expands, (b) a basic retrieval filter (top-20) only partially mitigates the decline . Crucially, end-to-end success is often bottlenecked by \emph{slot/value instantiation}: on ETID (Enterprise Tool Inventory Dataset, a synthetic dataset developed internally for this study),  \textbf{\autoref{fig:slot_filling_improve}} shows that improving slot filling produces large gains in overall success. Accordingly, our problem setting centers on reliable tool invocation under large inventories, where success depends on both correct tool selection and correct argument instantiation. 

Attempts to encode exhaustive tool and slot rules in lengthy global prompts are brittle, agents often fail to reliably follow extensive instructions, and maintaining cross-tool consistency becomes infeasible \cite{levy2024tasktokensimpactinput}. \textbf{\autoref{fig:example_enhance}} illustrates a representative tool-disambiguation failure that motivates JTPRO. In this example, two tools with overlapping descriptions, \texttt{get\_all\_countries} and \texttt{get\_countries\_list}, cause the baseline to mis-select the more generic tool for a request framed around investment analysis. After applying JTPRO, the tool descriptions are augmented with concise preference rules, specifying that \texttt{get\_all\_countries} should be used for general, non-investing requests, while \texttt{get\_countries\_list} should be preferred for investing or market-related queries. This resolves the ambiguity and yields the correct tool call, showing that targeted schema-level disambiguation can substantially improve tool selection in large tool inventories.

Prior work improves tool use via largely separate levers: model tuning, tuning-free prompting/documentation, retrieval-based tool filtering, and prompt/context optimization. Tuning-free prompting (CoT \cite{wei2023chainofthoughtpromptingelicitsreasoning}, ReAct \cite{yao2023reactsynergizingreasoningacting}) and documentation refinement (DRAFT \cite{qu2025explorationmasteryenablingllms}) avoid weight updates but typically treat global instructions and tool schemas as static; retrieval-based selection reduces overload and iterative variants refine retrievers with agent feedback \cite{xu2024toolretrievaliterative}, yet retrieval alone does not fix downstream argument/format errors when slot semantics remain unclear. Prompt optimization and context evolution methods MIPRO \cite{opsahl-ong-etal-2024-optimizing}; GEPA \cite{agrawal2025gepa}; AVATAR \cite{wu2024avatar}; Dynamic Cheatsheet \cite{suzgun2025dynamiccheatsheettesttimelearning}; ACE \cite{zhang2025agenticcontextengineeringevolving} improve instruction-level behavior, but do not \emph{jointly} adapt global decision rules and per-tool argument schemas at scale; similarly, \citet{wu-etal-2025-joint} refine prompts and tool descriptions but target efficiency rather than call-level tool/slot/value correctness under large domain tool stacks. JTPRO is best viewed as building on reflective optimization ideas and extending them to the \emph{joint} optimization of multiple agent operating components, global instructions and per-tool schema/argument descriptions for reliable tool calling.

\noindent Our core contributions are as follows:

- \textbf{Joint optimization of tool/slot-schema and global instructions (JTPRO).}
    We formulate \emph{joint} optimization of (i) the global instruction prompt $P$ and (ii) per-tool schema/argument descriptions $\{T_i\}$, targeting end-to-end invocation correctness (tool + slots + values) \emph{without} model fine-tuning. This is critical as tool-use failures are inherently \emph{coupled}: global policies depend on tool-local distinctions, and accurate slot/value instantiation relies on global conventions. Isolated optimization of $P$ or $\{T_i\}$ is insufficient to address these interdependent failure modes.

 \input{figures_camera_ready/tex/fig3_example_enhance}

- \textbf{Reflection-driven, localized edits with controlled growth.}
Inspired by reflection-augmented prompt engineering~\cite{agrawal2025gepa}, JTPRO diagnoses systematic rollout failures (tool confusion, missing constraints, and formatting/value errors) and issues targeted edits to both $P$ and the relevant tool/slot descriptions. To prevent context bloat, we \emph{globalize} recurring cross-tool slot semantics including date/time fields, numeric bounds, boolean parameters, sorting conventions, and currency/units into $P$, and replace redundant tool-local descriptions with short pointers to these shared rules. This reduces duplicated and potentially inconsistent schema text, while preserving tool-specific fields, exceptions, and disambiguation cues locally without merging or aliasing tools, which is important for real-world production systems; further details and examples are provided in Figures~\ref{fig:global_semantics_plot} and~\ref{fig:global_semantics_example}.

 - \textbf{Empirical evaluation under realistic constraints.} We benchmark JTPRO in both single- and multi-tool environments with variable argument structures, reporting Tool Selection Accuracy, Slot Filling Accuracy (conditional on tool correctness), and Overall Success Rate.
JTPRO demonstrates clear gains over strong baselines (such as baseline CoT, GEPA, and MIPRO) and further enhances retrieval-based pipelines by improving both retrieval and downstream slot filling.

%% file: figures_camera_ready/tex/fig1_jtpro_gains.tex
\begin{figure}[t]
    \centering
    \includegraphics[width=0.95\linewidth]{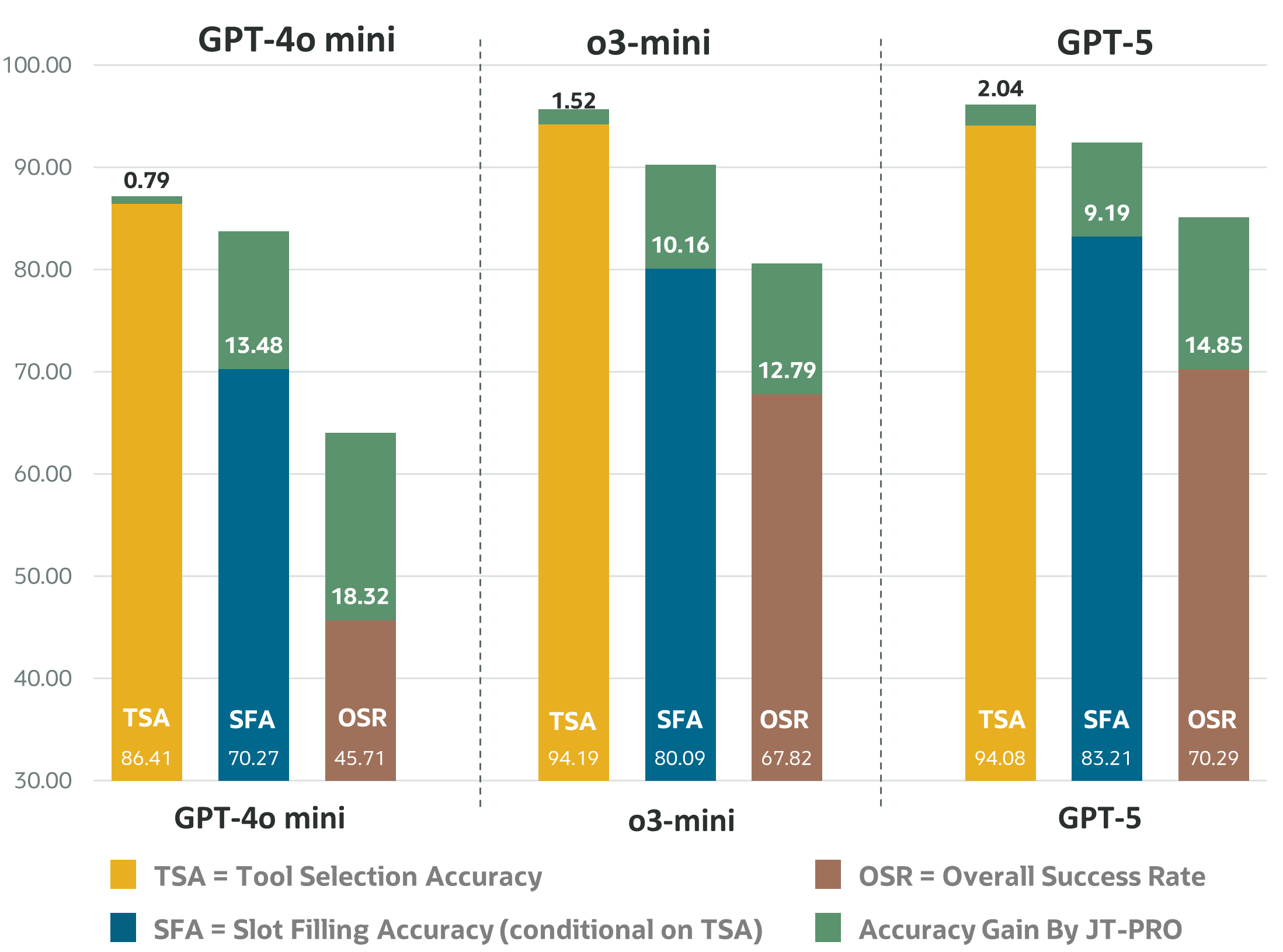}
    \caption{\textbf{Impact of slot-filling accuracy.}
    \emph{Slot filling drives end-to-end success:} On the \emph{Enterprise Tool-Inventory Dataset (ETID)} with complex schemas, we report TSA, SFA, and OSR; green overlays show absolute gains from JTPRO over baselines, highlighting that argument correctness is critical for OSR.}

    \label{fig:slot_filling_improve}
\end{figure}

%% file: figures_camera_ready/tex/fig2_rag_norag.tex
\begin{figure*}[t]
    \centering
    \includegraphics[width=0.8\linewidth]{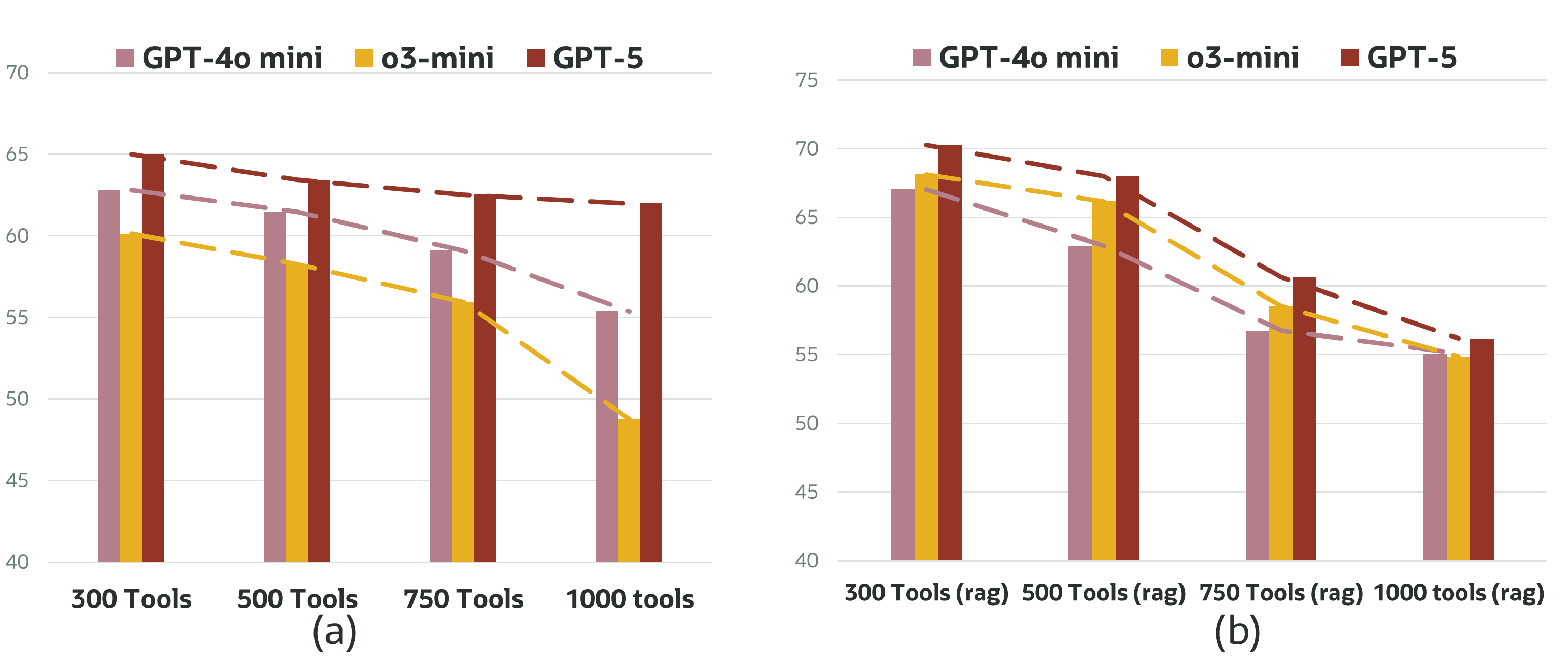}
    \caption{\textbf{Tool scaling failures and slot-filling impact.}
    \textbf{(a)} \emph{All tools in context:} On ToolACE with an augmented inventory, tool selection accuracy drops as the tool set grows ($300$ to $1000$), even for larger-context frontier models.
    \textbf{(b)} \emph{Top-$k$ retrieval:} A basic RAG with reranker stage (top-20) does not remove the drop, indicating residual tool disambiguation/argument issues.}

    \label{fig:tool_scaling_and_slot_filling}
\end{figure*}

%% file: figures_camera_ready/tex/fig3_example_enhance.tex
\begin{figure*}[htbp] 
    \centering
    \includegraphics[width=0.75\textwidth]{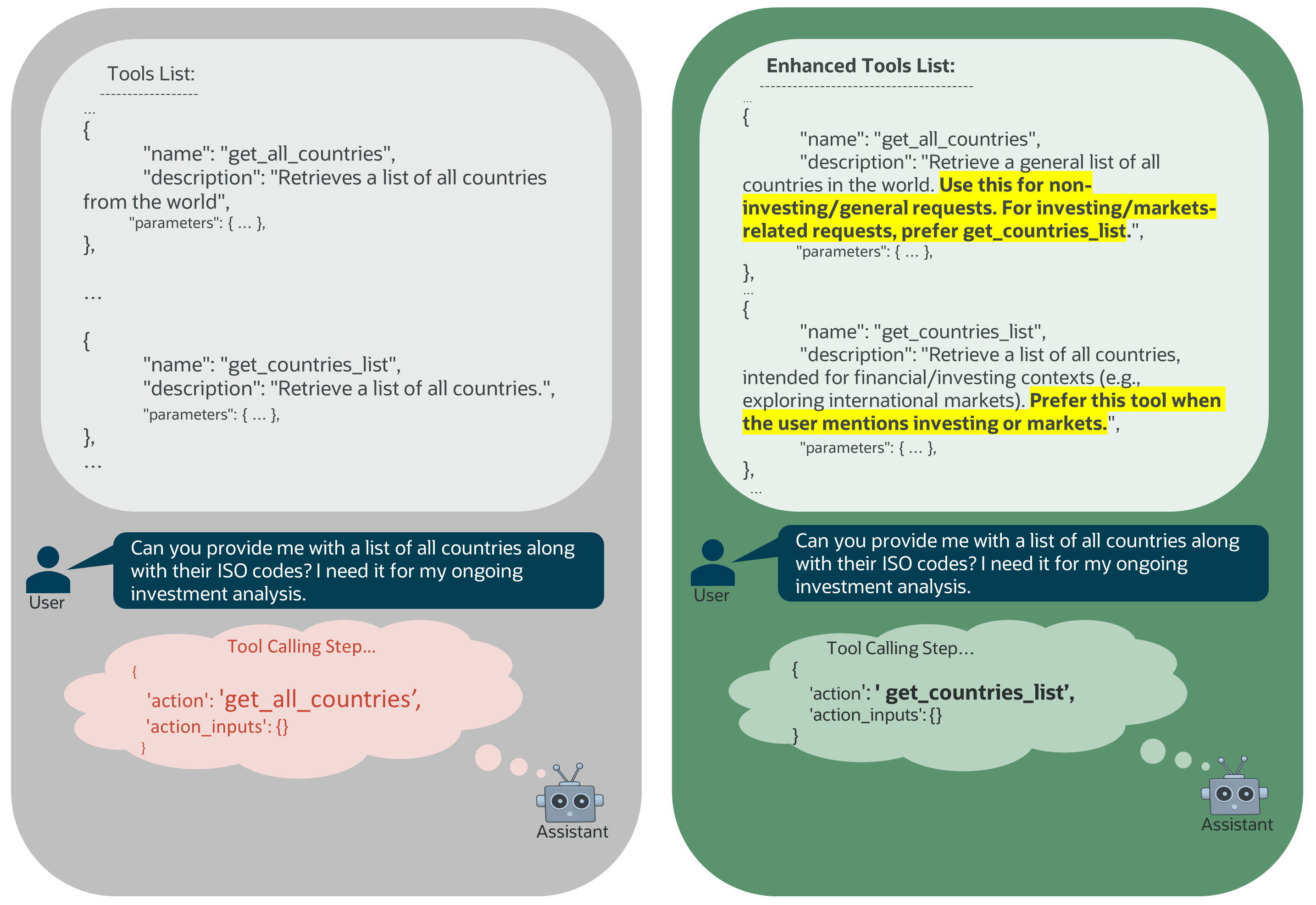} 
    \centering
    \caption{\textbf{Motivating context refinement with JTPRO, Tool disambiguation:} Baseline docs under-specify two similar tools, causing mis-selection; JTPRO adds brief per-tool decision rules (highlighted) to enable correct choice.}

    \label{fig:example_enhance}
\end{figure*}

%% file: sections/camera_ready/2-relatedwork.tex
\section{Related Work}
\label{sec:related}
Tool-use learning spans (i) tuning-based adaptation, (ii) tuning-free prompting and documentation refinement, (iii) retrieval-based tool selection, and (iv) prompt/context optimization. Most methods improve \emph{either} the global prompt \emph{or} tool specifications, but rarely their \emph{joint} co-adaptation under large, evolving tool inventories.

\textbf{Tuning-based tool learning.}
Model-tuning methods learn tool use by updating parameters or adding trainable modules from tool traces or preference/reward signals, including supervised fine-tuning \cite{qin2024toollearningfoundationmodels}, \cite{liu2025toolace}, contrastive objectives \cite{wu2024structureawarefinetuningcodepretrained}, reinforcement learning \cite{feng2025retoolreinforcementlearningstrategic, qian2025toolrlrewardtoollearning}, and tool-token embedding extensions \cite{Alazraki_2025}. While effective, these methods require retraining as the underlying tools/schemas evolve.

\input{figures_camera_ready/tex/fig_4_algorithm}

\textbf{Tuning-free prompting and documentation refinement.}
Prompting approaches like CoT and ReAct and agentic planners like RestGPT \cite{song2023restgptconnectinglargelanguage} and HuggingGPT \cite{shen2023hugginggptsolvingaitasks} elicit multi-step reasoning without weight updates, but usually treat instructions and tool schemas as static. DRAFT \cite{qu2025explorationmasteryenablingllms} improves per-tool documentation via trial-and-error, yet does not optimize global instruction policies or multi-tool interactions mediated by shared prompt rules.

\textbf{Retriever-based tool selection.}
Retriever-based pipelines filter candidates via lexical/dense retrieval and specialized rerankers e.g., CRAFT \cite{yuan2024craft}, ToolRerank \cite{zheng2024toolrerank}, COLT \cite{Qu_2024}, improving scalability but not resolving argument/format errors when slot semantics are unclear. Iterative retrieval refinement with agent feedback \cite{xu2024toolretrievaliterative, ag2025aligning, pattnayak2025hybrid} reduces retriever--agent mismatch, but typically leaves the agent’s instruction layer largely unchanged.

\textbf{Tool-using agents, tool construction, and prompt optimization.}
Toolformer \citep{schick2023toolformer}, ReAct \citep{yao2023react}, and ReWOO \citep{xu2023rewoo} integrate tool calls into reasoning traces; DSPy \citep{khattab2024dspy} and AutoPDL \citep{2025arXiv250404365S} support declarative tool programs but assume static prompts/schemas. Other work constructs tools (TOOLMAKER \citep{wolflein-etal-2025-llm}), optimizes tool-use prompts (AvaTaR \citep{wu2024avatar}), calibrates tool use (CITI~\citep{10.1609/aaai.v39i22.34573}, PROBECAL~\citep{liu-etal-2024-uncertainty}), or improves tool policies via SFT/RL \citep{sullivan-etal-2025-procedural}.
Separately, self-refinement and prompt optimization \citep{madaan2023, shin-etal-2020-autoprompt, lester-etal-2021-power, pryzant-etal-2023-automatic, yuksekgonul2025optimizing, singh2026mtoscpathllmslost, zheng2026diffumaskdiffusionlanguagemodel} and evolutionary search EvoPrompt \citep{guo2024connecting} automate instruction improvement; MIPRO \cite{opsahl-ong-etal-2024-optimizing} optimizes module prompts and demonstrations, while GEPA \citep{agrawal2025gepa} uses reflection over trajectories with Pareto selection, and AVATAR \citep{wu2024avatar} applies contrastive feedback. Dynamic Cheatsheet \cite{suzgun2025dynamiccheatsheettesttimelearning} and ACE \cite{zhang2025agenticcontextengineeringevolving} motivate maintaining an evolving, curated context, but focus on strategy/memory rather than tool/argument schema co-adaptation.

\textbf{Reflective textual feedback for prompt/text optimization.}
Recent work moves beyond scalar rewards (e.g., accuracy) by using \emph{rich textual critiques} as the optimization signal, treating LLMs as optimizers that make targeted, gradient-like edits in text space. Maestro \citep{maestro_arxiv} places prompt optimization in a broader system loop, jointly updating agent graphs and configurations (including prompts) from reflective feedback over execution traces such as constraint violations and looping, improving reliability and sample efficiency. Feedback Descent \citep{feedbackdescent_arxiv} extends this view to an open-ended framework that uses pairwise comparisons, textual rationales, and accumulated feedback to iteratively revise prompts and other text artifacts. These approaches are complementary to tool-use settings: they show the value of structured, interpretable feedback for inference-time refinement, but do not directly address joint co-adaptation of \emph{global instruction policies} and \emph{per-tool argument/schema descriptions} in large tool inventories.

\textbf{Distinction.}
In contrast to prior work that optimizes prompts or tool documentation separately, JTPRO jointly updates global instructions $P$ and per-tool \emph{tool/argument} schema descriptions $\{T_i\}$ using rollout-driven reflection, targeting call-level correctness (tool, slots, values) without model fine-tuning. JTPRO also reduces redundancy by abstracting shared slot conventions globally while preserving tool-specific details locally, leading to improved results in retrieval-based pipelines.

%% file: figures_camera_ready/tex/fig_4_algorithm.tex
\begin{figure*}[htbp] 
    \centering
    \includegraphics[width=0.9\textwidth]{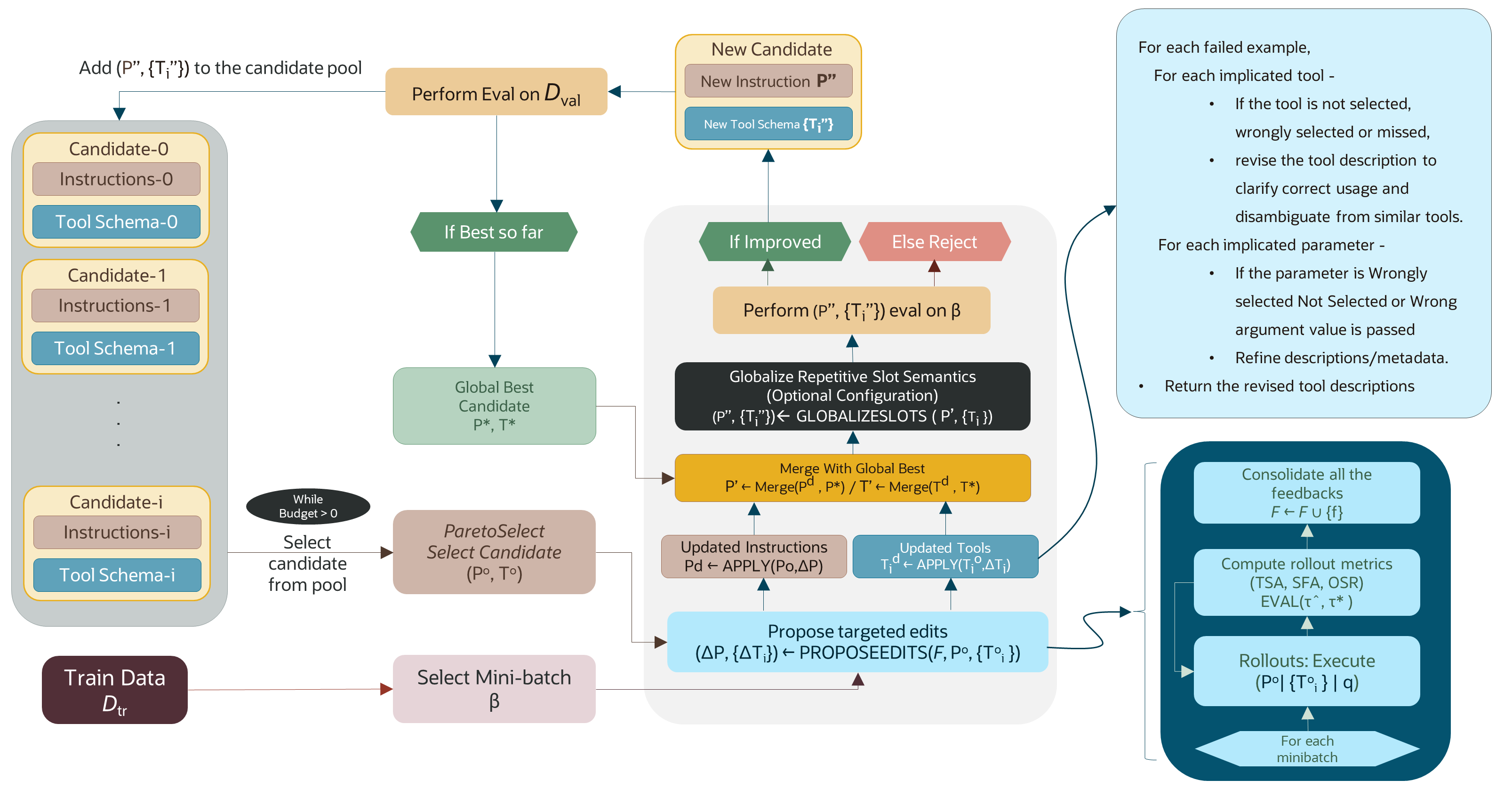} 
    \caption{\textbf{JTPRO optimization loop (block-diagram view).} 
JTPRO maintains a pool of candidate contexts (global instructions $P$ and tool schemas $\{T_i\}$) and repeatedly (i) selects a candidate via Pareto-based sampling, (ii) runs minibatch rollouts on $\mathcal{D}_{tr}$ to compute tool-use metrics (\textsc{TSA}, \textsc{SFA}, \textsc{OSR}) and aggregate error feedback, and (iii) proposes localized edits to both $P$ and the implicated tool schemas. The edited instructions are merged with the current global-best $(P^\star,\{T_i^\star\})$, followed by optional globalization of repetitive slot semantics to avoid duplicated cross-tool parameter rules. Candidates that improve minibatch performance are validated on $\mathcal{D}_{val}$; improved candidates are added back to the pool, and the global best $(P^\star,\{T_i^\star\})$ is updated when a new highest validation score is observed.}
    \label{fig:figjointtoolprompt}
\end{figure*}

%% file: sections/camera_ready/3-probstatement.tex
\section{Problem Statement}

We consider an LLM agent with access to a set of $N$ external tools (APIs/functions)
$\{T_1,\dots,T_N\}$. Each tool $T_i$ is specified by a schema/documentation entry
describing its functionality and expected parameters (slots). Given a user query $Q$,
the agent must produce an answer $A$, potentially by issuing one or more tool calls with
structured arguments. The agent is guided by a global instruction prompt $P$ and the
collection of tool schemas $\{T_i\}_{i=1}^N$.

For a query $Q$, the LLM is invoked with context
\begin{equation}
C(P,T,Q)\;=\;P\;\big|\;T_1\;\big|\;\cdots\;\big|\;T_N\;\big|\;Q,
\label{eq:context-simple}
\end{equation}
and produces a tool-call trace $\hat{\tau}=\hat{\tau}(P,T,Q)$.

Our objective is to optimize the textual content of $P$ and $\{T_i\}$ to maximize tool-use
performance \emph{without} model fine-tuning. Because tool identities and interfaces are
typically fixed in production, we do \emph{not} merge or alias tools. Instead, we allow edits to
$P$ and each $T_i$, and we \emph{globalize} recurring slot conventions (e.g., date/time formats,
inclusive/exclusive bounds, currency/units) by lifting duplicated per-tool guidance into $P$.

We evaluate \textbf{call-level correctness}: correct tool selection and correct slot/value
instantiation, summarized by \textbf{Tool Selection Accuracy}, \textbf{Slot Filling Accuracy}
(conditional on correct tool), and \textbf{Overall Success Rate} (correct tool $+$ correct slots
$+$ correct values). This emphasis matches deployments where executing tools and validating
response-level correctness may be infeasible due to security, access control, rate limits, or
non-deterministic backends.

Given a dataset $\mathcal{D}=\{(Q_j,\tau_j)\}_{j=1}^M$ with gold traces $\tau$, we optimize only two variables the global instructions $P$ and tool
descriptions $T$ to maximize expected call-level correctness:
\begin{equation}
(P^\star,T^\star)
=\arg\min_{P,T}\;
\mathbb{E}_{(Q,\tau)\sim\mathcal{D}}
\Big[
\mathcal{L}\!\big(\hat{\tau}(P,T,Q),\tau)
\Big].
\label{eq:jtpo-objective-simple}
\end{equation}

The loss function can be defined using tool selection, slot filling, and overall success:
\begin{equation}
\begin{split}
\mathcal{L}(\hat{\tau},\tau)
=
\lambda_{\textsc{tsa}}\,(1-\mathbb{I}[\hat{t}=t]) \\
+
\lambda_{\textsc{sfa}}\, 
\mathbb{I}[\hat{t}=t]\,
\big(1-\mathrm{Rec}(\hat{a},a)\big) \\
+
\lambda_{\textsc{osr}}\,
\big(1-\mathbb{I}[\hat{t}=t \wedge \hat{a}=a]\big),
\label{eq:jtpo-loss-simple}
\end{split}
\end{equation}
where $\hat{t}$ and $t$ are the predicted and gold tool identifiers, $\mathbb{I}[\cdot]$ is the indicator function, $\hat{\mathbf{a}}$ and $\mathbf{a}$ are the predicted and gold argument structures,  $\mathrm{Rec}(\hat{a},a)$ are slot/value recall conditional on $\hat{t}=t$, and $\lambda_{\textsc{tsa}},\lambda_{\textsc{sfa}},\lambda_{\textsc{osr}}$ are nonnegative loss weights.

%% file: sections/camera_ready/4-technique.tex
\section{Technique}

We present \textbf{Joint Tool--Prompt Reflective Optimization (JTPRO)}, a weight-free, context-level optimizer that iteratively updates (i) global agent instructions $P$ and (ii) per-tool schemas $\{T_i\}_{i=1}^N$ from labeled tool-call traces. Algorithm~\ref{alg:JTPRO} summarizes the loop.

\paragraph{Setup and objective}
For each query $q$, the agent runs under $C(q)=P \,|\, T_1 \,|\, \cdots \,|\, T_N \,|\, q$ and produces a predicted trace $\hat{\tau}$. Given gold traces $\tau^\star$, JTPRO edits $P$ and $\{T_i\}$ to improve \textsc{TSA}, \textsc{SFA} (conditional on correct tool), and \textsc{OSR} (correct tool + correct slots + correct values).

\paragraph{Candidate selection (Pareto)}
JTPRO maintains a pool $\mathcal{C}$ of candidate contexts and uses GEPA-style Pareto selection: retain candidates that achieve the best score on at least one training instance, prune strictly dominated candidates, then sample a starting candidate with probability biased toward those that win on more instances.

\paragraph{Rollouts, diagnostics, and localized edits}
On a minibatch $\mathcal{B}\subset\mathcal{D}_{tr}$, we compute rollout metrics and extract structured failure signals $\mathcal{F}$ via \textsc{Diagnose}$(\hat{\tau},\tau^\star)$ (e.g., tool confusions, missing required slots, formatting/value violations). A reflector proposes targeted edits $(\Delta P,\{\Delta T_i\})\leftarrow \textsc{ProposeEdits}(\mathcal{F},P^o,\{T_i^o\})$, which are applied to produce a draft context $P^d$ and $\{T_i^d\}$. Edits are localized to the implicated global rules and tool/slot descriptions.

\begin{algorithm}
\small
\DontPrintSemicolon
\caption{JTPRO: Reflective Schema--Instruction Co-Optimization with Slot-Semantics Globalization}
\label{alg:JTPRO}

\textbf{Input:} initial global instructions $P^{(0)}$, initial tool schemas $\{T_i^{(0)}\}_{i=1}^N$, labeled training set $\mathcal{D}_{tr}=\{(q,\tau^\star)\}$, labeled validation set $\mathcal{D}_{val}$, max iterations $I$, batch size $B$ \\
\textbf{Output:} optimized global instructions $P^\star$, optimized tool schemas $\{T_i^\star\}_{i=1}^N$

\BlankLine
Initialize $P \leftarrow P^{(0)}$ and $T_i \leftarrow T_i^{(0)}$ for all $i \in \{1,\dots,N\}$\;
Initialize best context $C^\star \leftarrow (P,\{T_i\}_{i=1}^N)$ and best validation score $s^\star \leftarrow -\infty$\;
Initialize pool $\mathcal{C} \leftarrow \{(P,\{T_i\}_{i=1}^N)\}$ \tcp*{candidate contexts}

\For(\tcp*[f]{main optimization loop}){$t \leftarrow 1$ \KwTo $I$}{
    Sample a minibatch $\mathcal{B} \subset \mathcal{D}_{tr}$ of size $B$\;
    $(P^o,\{T_i^o\}_{i=1}^N) \leftarrow \textsc{ParetoSelect}(\mathcal{C})$\;
    $s^o_{\mathcal{B}} \leftarrow \textsc{Score}((P^o,\{T_i^o\}_{i=1}^N), \mathcal{B})$\;
    Initialize aggregated feedback $\mathcal{F} \leftarrow \emptyset$\;

    \ForEach{$(q,\tau^\star) \in \mathcal{B}$}{
        Construct context $C(q) \leftarrow P^o \,|\, T_1^o \,|\, \cdots \,|\, T_N^o \,|\, q$\;
        Run agent to obtain predicted trace $\hat{\tau} \leftarrow \textsc{Agent}(C(q))$\;
        Compute rollout metrics $(\textsc{TSA},\textsc{SFA},\textsc{OSR}) \leftarrow \textsc{Eval}(\hat{\tau},\tau^\star)$\;
        Extract error signals $f \leftarrow \textsc{Diagnose}(\hat{\tau},\tau^\star)$\;
        $\mathcal{F} \leftarrow \mathcal{F} \cup \{f\}$\;
    }

    $(\Delta P,\{\Delta T_i\}_{i=1}^N) \leftarrow \textsc{ProposeEdits}(\mathcal{F}, P^o, \{T_i^o\}_{i=1}^N)$\;
    $P^d \leftarrow \textsc{Apply}(P^o, \Delta P)$\;
    $T_i^d \leftarrow \textsc{Apply}(T_i^o, \Delta T_i)$ for all $i \in \{1,\dots,N\}$\;

    $P' \leftarrow \textsc{Merge}(P^d, P^\star)$\;
    $T_i' \leftarrow \textsc{Merge}(T_i^d, T_i^\star)$ for all $i \in \{1,\dots,N\}$\;

    $(P'',\{T_i''\}_{i=1}^N) \leftarrow \textsc{GlobalizeSlots}(P', \{T_i'\}_{i=1}^N)$\;

    $s''_{\mathcal{B}} \leftarrow \textsc{Score}((P'',\{T_i''\}_{i=1}^N), \mathcal{B})$\;
    \If{$s''_{\mathcal{B}} > s^o_{\mathcal{B}}$}{
        $s''_{\mathrm{val}} \leftarrow \textsc{Score}((P'',\{T_i''\}_{i=1}^N), \mathcal{D}_{val})$\;
        \If{$s''_{\mathrm{val}} \ge s^\star$}{
            $\mathcal{C} \leftarrow \textsc{AddToPool}(\mathcal{C}, (P'',\{T_i''\}_{i=1}^N), K)$\;
            \If{$s''_{\mathrm{val}} > s^\star$}{
                $C^\star \leftarrow (P'',\{T_i''\}_{i=1}^N)$\;
                $(P^\star,\{T_i^\star\}_{i=1}^N) \leftarrow (P'',\{T_i''\}_{i=1}^N)$\;
                $s^\star \leftarrow s''_{\mathrm{val}}$\;
            }
        }
    }
}
\Return{$C^\star$ as $(P^\star,\{T_i^\star\}_{i=1}^N)$}\;
\end{algorithm}

\paragraph{Merge-with-best for incremental tool adaptation}
JTPRO tracks a validation-best context $C^\star=(P^\star,\{T_i^\star\})$. After editing, we merge $P^d$ with $P^\star$ using \textsc{Merge}$(P^d,P^\star)$ to form $P'$, and \textsc{Merge}$(T^d,T^\star)$ to form $T'$, implementing a ``growing playbook'' that preserves cross-cutting rules while adding new, rollout-driven guidance. This accumulation also supports incremental toolset expansion: when new $T_i$ are appended, stable global conventions remain intact and new tool-triggered rules are integrated without re-optimizing from scratch.

\paragraph{Globalizing repetitive slot semantics}
To reduce duplicated schema text, JTPRO applies \textsc{GlobalizeSlots}$(P',\{T_i'\})\mapsto(P'',\{T_i''\})$, which identifies recurring cross-tool slot conventions and lifts them into named rules in $P''$ while replacing redundant tool-local descriptions with short pointers to those rules. Figure~\ref{fig:global_semantics_plot} motivates this step: in ETID, a small number of slot families, especially identifiers and date/time fields, but also numeric bounds, boolean flags, sorting parameters, and currency/unit fields---recur across many tools (up to 77/124), producing substantial repetition in per-tool schemas. Figure~\ref{fig:global_semantics_example} illustrates the resulting two-level organization: shared fields such as \texttt{startDate}, \texttt{endDate}, \texttt{rangeMinimum}, \texttt{rangeMaximum}, and their inclusive flags point to global rules (e.g., \emph{DateTime Fields}, \emph{Numeric Bounds}, and \emph{Boolean Parameters}), while tool-specific fields and local overrides remain in $T_i''$. This improves slot filling by enforcing consistent semantics across tools, reducing duplicated and potentially conflicting wording, and preserving schema space for tool-specific exceptions and disambiguation rules.

\paragraph{Acceptance and pool update}
We score $(P'',\{T_i''\})$ on $\mathcal{B}$ and, if improved, evaluate on $\mathcal{D}_{val}$. Improved candidates are added to $\mathcal{C}$ (bounded size $K$), and if a candidate is best on validation we update $C^\star$ accordingly.

\paragraph{Summary}
JTPRO combines Pareto-selected candidate search, reflection-driven localized edits to $P$ and $\{T_i\}$, and globalization of shared slot semantics to improve both tool selection and argument correctness in large tool inventories.

%% file: sections/camera_ready/5-evalsetup.tex
\section{Datasets and Evaluation}
\label{sec:data_eval}

\subsection{Datasets}
We evaluate JTPRO on three complementary benchmarks that stress different failure modes in tool-using agents: (i) complex, domain-specific slot filling with a moderate tool inventory, (ii) tool selection under toolset scaling, and (iii) a \emph{multi-tool calling} setting where a single query may require invoking multiple tools in parallel and correctly instantiating arguments at each step. Our benchmark choices are matched to JTPRO’s core setting: reusable tool schemas and stable train/validation/test distributions that permit transferable prompt and schema refinement, particularly in large tool inventories where tool selection and schema-constrained argument filling are the dominant bottlenecks. We discuss benchmark suitability, including the omission of other popular datasets, as well as the current scope of our evaluation with respect to sequential tool dependencies, in Appendix~\ref{app:benchmark_scope}.

\paragraph{Enterprise Tool-Inventory Dataset (ETID).}
ETID is a domain-specific tool-calling dataset targeting \emph{argument correctness} under complex schemas. It contains $124$ tools with $3.4$ parameters on average (max $12$) and $\sim$13 labeled examples per tool (min $10$). We evaluate both an \emph{all-tools} setting and \emph{value-stream} subsets. For data efficiency, we use intent-aligned regimes \texttt{Train-$N$ex} where each tool contributes $N$ train and $N$ validation examples (total $124\times N$), reporting \texttt{Train-1ex/2ex/4ex}. The test set is fixed at $404$ queries.

\paragraph{ToolACE (tool scaling).}
ToolACE evaluates performance degradation as the tool universe expands. We use fixed splits (Train $=199$, Validation $=76$, Test $=121$) and augment the tool inventory to create \texttt{ToolACE-300/500/750/1000} variants.

\paragraph{SEAL-Tools (parallel multi-tool calling).}
SEAL-Tools~\citep{wu2024sealtools} benchmarks \emph{parallel} multi-tool calling across diverse domains. We use a curated \texttt{multiple-overlap} subset with $1{,}138$ tools and $2{,}743$ arguments, split into Train $=600$, Validation $=100$, Test $=100$. Each query requires $3.2$ parallel tool calls on average (typically $3$), with $5.8$ arguments filled per query, stressing joint multi-tool selection and argument filling.

  We use a curated \texttt{multiple-overlap} subset containing $1{,}138$ tools with $2{,}743$ arguments. The split is Train $=600$, Validation $=100$, Test $=100$ examples. Each query requires an average of $3.2$ parallel tool calls ($77\%$ require exactly $3$ tools) with $5.8$ arguments filled per query. Tool coverage overlap ensures training tools appear in evaluation splits. This setting isolates the challenge of \emph{joint} tool selection and slot filling at scale, where models must correctly identify multiple tools and fill all arguments for each.

\subsection{Evaluation Metrics}
Following prior tool-use evaluations, we measure call-level correctness rather than answer accuracy.
Specifically, we report: 

\begin{itemize}
    \item \textbf{Tool Selection Accuracy (TSA)}: fraction of queries for which the agent chose the correct tool(s) required (including choosing none if no tool needed).
    \item \textbf{Slot Filling Accuracy (SFA)}: recall of correct slot/value assignments \emph{conditional on correct tool selection}.
    \item \textbf{Overall Success Rate (OSR)}: (correct tool + correct slots + correct values).
\end{itemize}
This evaluation reflects practical deployments where executing the true tool backend may be infeasible (e.g., security constraints, access controls, rate limits, or non-deterministic systems), so correctness must be assessed at the tool-call level.

\label{sec:dataset_stats}

  \begin{table}[t]
  \centering
  \small
  \setlength{\tabcolsep}{6pt}
  \begin{tabular}{lrrrrr}
  \toprule
  \multirow{2}{*}{\textbf{Dataset}} & \multirow{2}{*}{\textbf{\#Tools}} &
  \multicolumn{2}{c}{\textbf{Total Args}} &
  \multicolumn{2}{c}{\textbf{Required Args}} \\
  \cmidrule(lr){3-4}\cmidrule(lr){5-6}
   &  & \textbf{Avg} & \textbf{Max} & \textbf{Avg} & \textbf{Max} \\
  \midrule
  ETID          & 124  & 3.4  & 12 & 0.81 & 6 \\
  \midrule
  ToolACE-300   & 336  & 2.05 & 14 & 1.20 & 6 \\
  ToolACE-500   & 536  & 2.14 & 17 & 1.20 & 7 \\
  ToolACE-750   & 786  & 2.17 & 23 & 1.21 & 7 \\
  ToolACE-1000  & 1036 & 2.10 & 23 & 1.21 & 7 \\
  \midrule
  SEAL-Tools    & 1138 & 2.41 & 8  & 1.60 & 5 \\
  \bottomrule
  \end{tabular}
  \caption{Dataset statistics. ``\#Tools'' denotes the size of the tool universe available at inference time. ``Total Args'' counts all schema parameters per tool, and ``Required Args'' counts mandatory parameters.}
  \label{tab:dataset_stats}
  \end{table}

%% file: sections/camera_ready/6-resultsanalysis.tex
\section{Results and Analysis}
\begin{table*}[t]
\centering
\setlength{\tabcolsep}{5pt}
\begin{tabular}{l c | ccc | ccc | ccc}
\toprule
\multirow{2}{*}{Model} & \multirow{2}{*}{\#Tools} &
\multicolumn{3}{c|}{\textsc{TSA} (\%)} &
\multicolumn{3}{c|}{\textsc{SFA} (\%) $\uparrow \,|\, \textsc{TSA}$} &
\multicolumn{3}{c}{\textsc{OSR} (\%)} \\
& & Base & GEPA & JTPRO & Base & GEPA & JTPRO & Base & GEPA & JTPRO \\
\midrule
GPT-4o mini & 500  & 63.832 & 73.33 & \textbf{75.25} & 86.996 & 85.27 & \textbf{88.12} & 60.00 & 61.98 & \textbf{69.42} \\
GPT-4o mini & 1000 & 61.115 & 73.39 & \textbf{75.13} & 86.67  & 83.36 & \textbf{83.59} & 58.18 & 60.33 & \textbf{63.64} \\
\midrule
o3-mini & 500  & 70.78  & 73.33 & \textbf{76.19} & 84.994 & 85.27 & \textbf{88.52} & 59.454 & 61.98 & \textbf{65.29} \\
o3-mini & 1000 & 58.916 & 70.11 & \textbf{71.48} & 85.036 & 86.59 & \textbf{87.46} & 51.272 & 58.68 & \textbf{64.46} \\
\midrule
GPT-5 & 500  & 73.02  & 77.17 & \textbf{82.28} & 84.785 & 85.75 & \textbf{90.00} & 62.73  & 66.12 & \textbf{74.38} \\
GPT-5 & 1000 & 67.658 & 75.13 & \textbf{78.72} & 87.35  & 86.40 & \textbf{89.26} & 62.366 & 67.77 & \textbf{73.55} \\
\bottomrule
\end{tabular}
\caption{\textbf{ToolACE results under tool-universe scaling (500 vs.\ 1000 tools).} JTPRO achieves the strongest end-to-end performance (\textsc{OSR}) by jointly improving tool selection (\textsc{TSA}) and argument correctness (\textsc{SFA}), with the largest gains appearing in the 1000-tool regime where tool confusions are most frequent.}
\label{tab:toolace_main}
\end{table*}

\subsection{ToolACE: Scaling the Tool Universe}
Table~\ref{tab:toolace_main} reports Tool Selection Accuracy (\textsc{TSA}), Slot Filling Accuracy (\textsc{SFA}; conditional on correct tool), and Overall Success Rate (\textsc{OSR}; correct tool + correct slots + correct values) for ToolACE with 500 and 1000 tools.

As the tool inventory grows, baseline performance drops primarily in \textsc{TSA}, which cascades to lower \textsc{OSR} even when \textsc{SFA} remains high. GEPA improves \textsc{TSA} in most settings, but gains in \textsc{OSR} are limited because failures often stem from tool-specific disambiguation and argument constraints that global instruction refinement alone cannot resolve.

JTPRO consistently achieves the highest \textsc{TSA} and \textsc{OSR} across all models and tool counts. Gains are especially pronounced in the 1000-tool setting (e.g., +13.2 \textsc{OSR} points for \texttt{o3-mini} over baseline). While \textsc{SFA} is already strong, JTPRO further boosts end-to-end success by reducing tool confusions and encoding missing slot/value conventions. These results show that, on ToolACE, \textsc{OSR} improvements are primarily driven by better tool selection, emphasizing that accurate \textsc{TSA} is critical for downstream argument correctness.

\begin{table*}[t]
\centering
\setlength{\tabcolsep}{5pt}
\begin{tabular}{l c | ccc | ccc | ccc}
\toprule
\multirow{2}{*}{Model} & \multirow{2}{*}{Train} &
\multicolumn{3}{c|}{\textsc{TSA} (\%)} &
\multicolumn{3}{c|}{\textsc{SFA} (\%) $\uparrow \,|\, \textsc{TSA}$} &
\multicolumn{3}{c}{\textsc{OSR} (\%)} \\
& & Base & GEPA & JTPRO & Base & GEPA & JTPRO & Base & GEPA & JTPRO \\
\midrule
GPT-4o mini & Train-1ex & 85.91 & 86.23 & 83.14 & 69.81 & 78.71 & \textbf{82.72} & 44.80 & 50.19    & \textbf{60.15} \\
GPT-4o mini & Train-2ex & 86.36 & 85.32 & 88.27 & 70.90 & 77.12    & \textbf{83.37} & 45.79 & 52.7    & \textbf{65.10} \\
GPT-4o mini & Train-4ex & 86.96 & 88.38 & 87.18 & 70.09 & 74.93    & \textbf{85.16} & 46.53 & 54.34    & \textbf{66.83} \\
\midrule
o3-mini & Train-1ex & 94.00 & 95.08 & 95.78 & 80.48 & 83.73 & \textbf{89.77} & 68.81 & 75.00 & \textbf{79.46} \\
o3-mini & Train-2ex & 94.15 & 94.40 & 95.40 & 80.20 & 85.30 & \textbf{90.01} & 67.33 & 73.02 & \textbf{79.70} \\
o3-mini & Train-4ex & 94.41 & 94.90 & 95.95 & 79.60 & 87.455& \textbf{90.98} & 67.325& 77.725& \textbf{82.67} \\
\midrule
GPT-5 & Train-1ex & 94.06 & 97.27 & 95.76 & 82.45 & 90.21 & \textbf{92.15} & 68.81 & 80.20 & \textbf{84.65} \\
GPT-5 & Train-2ex & 94.03 & 97.91 & 95.80 & 83.90 & 89.89 & \textbf{92.77} & 71.53 & 79.18 & \textbf{85.64} \\
GPT-5 & Train-4ex & 94.16 & 98.47 & 96.81 & 83.29 & 88.80 & \textbf{92.30} & 70.54 & 80.69 & \textbf{85.15} \\
\bottomrule
\end{tabular}
\caption{\textbf{ETID results under low-supervision training regimes.} JTPRO yields the most consistent \textsc{OSR} improvements, indicating that improved argument semantics are crucial for complex enterprise schemas.}
\label{tab:etid_main}
\end{table*}

\begin{table*}[t]
\centering
\setlength{\tabcolsep}{5pt}
\begin{tabular}{l | ccc | ccc | ccc}
\toprule
\multirow{2}{*}{Model} &
\multicolumn{3}{c|}{\textsc{TSA} (\%)} &
\multicolumn{3}{c|}{\textsc{SFA} (\%) $\uparrow \,|\, \textsc{TSA}$} &
\multicolumn{3}{c}{\textsc{OSR} (\%)} \\
 & Base & GEPA & JTPRO & Base & GEPA & JTPRO & Base & GEPA & JTPRO \\
\midrule
GPT-4o & 81	& 82.11	& 82.3 & 40.4 & 43.51 & \textbf{53.51} & 23 & 24.51 & \textbf{27.5} \\
\midrule
o3-mini & 82.2 & 82.1 & 83.9  & 52.3 & 56.9 & \textbf{60.1} & 26.3 & 27.5 & \textbf{30.1} \\
\midrule
GPT-5 & 84.5 & 85.2 & 86.5  & 56.3 & 61.3 & \textbf{65.2} & 28.8 & 31.1 & \textbf{33.6} \\
\bottomrule
\end{tabular}
\caption{\textbf{SEAL-Tools results (multi-tool calling).} JTPRO consistently improves \textsc{SFA} and \textsc{OSR} across all models, while keeping \textsc{TSA} stable or slightly improved.}
\label{tab:sealtools_results}
\end{table*}

\subsection{ETID: Complex Slot Filling with Moderate Tool Counts}
We evaluate on the \emph{Enterprise Tool-Inventory Dataset (ETID)}, which features complex multi-argument schemas (avg.\ 3.4 parameters/intent; max 12) and measures correctness at the \emph{call level} (tool + slots + values). Table~\ref{tab:etid_main} reports results under low-supervision regimes (Train-1/2/4 examples per intent; fixed test set of 404 queries).

Two trends stand out. First, \textbf{slot/value correctness is the main bottleneck}. Baseline \textsc{TSA} is high (85--94\%), yet \textsc{OSR} remains much lower, showing that \textsc{SFA} errors dominate once the correct tool is chosen. JTPRO addresses this directly, improving \textsc{SFA} and boosting \textsc{OSR}—e.g., for \texttt{GPT-4o mini}, \textsc{OSR} rises from 44.8$\rightarrow$60.15 (+15.35) in Train-1ex and 46.53$\rightarrow$66.83 (+20.30) in Train-4ex, despite similar \textsc{TSA}.

Second, \textbf{JTPRO delivers robust gains across models and training regimes}. For \texttt{gpto3-mini}, \textsc{OSR} improves over both baseline and GEPA in all regimes, with larger gains as supervision increases (Train-4ex: 67.33$\rightarrow$82.67). For \texttt{GPT-5}, GEPA raises \textsc{TSA}, but JTPRO achieves the highest \textsc{OSR} by combining strong tool selection with higher \textsc{SFA} (Train-2ex: \textsc{SFA} 92.77, \textsc{OSR} 85.64). Overall, ETID shows that optimizing tool selection alone is insufficient; joint refinement of instructions and tool/slot descriptions is necessary to convert high \textsc{TSA} into end-to-end success.

\subsection{SEAL-Tools: Multi-Tool Calling}
We evaluate on the SEAL-Tools multi-tool subset, which contains 1{,}138 tools and requires an average of 3.2 parallel tool calls per query. This makes both tool disambiguation and argument instantiation challenging, since success depends on selecting the correct tools and correctly filling their arguments across multiple calls.


Table~\ref{tab:sealtools_results} shows that JTPRO consistently improves \textsc{SFA} and \textsc{OSR} across all three models, while keeping \textsc{TSA} stable or slightly improving it. For \textbf{GPT-4o}, \textsc{TSA} changes only slightly from 81.0 to 82.3, while \textsc{SFA} rises from 40.4 to 53.51 and \textsc{OSR} from 23.0 to 27.5. For \textbf{o3-mini}, \textsc{TSA} improves from 82.2 to 83.9, with \textsc{SFA} increasing from 52.3 to 60.1 and \textsc{OSR} from 26.3 to 30.1. For \textbf{GPT-5}, \textsc{TSA} increases from 84.5 to 86.5, alongside gains in \textsc{SFA} from 56.3 to 65.2 and \textsc{OSR} from 28.8 to 33.6.

Overall, the results reinforce our central finding: in large, schema-rich multi-tool settings, improving tool selection alone is not enough. Joint refinement of agent instructions and per-tool schema descriptions is needed to translate strong \textsc{TSA} into better end-to-end success.                                                                                            

\subsection{Instance-Level Slot Corrections and Tool Disambiguation}
JTPRO improves slot/value instantiation at the example level while also making semantically similar tools easier to distinguish. Figure~\ref{fig:toolace500_gpt5_example_slot_impr} shows that for GPT-5 on \textbf{ToolACE-500}, JTPRO corrects previously incorrect slot/value assignments on \textbf{26 of 121} test examples (\textbf{21.48\%}), while on \textbf{ETID} (Figure~\ref{fig:etid_gpt5_example_slot_impr}) it improves slot correctness on \textbf{94 of 403} examples (\textbf{23.33\%}). Because these gains are distributed across many test instances, they indicate that JTPRO’s improvements are broad rather than driven by a small number of outliers.

Another key reason of these gains is improved tool description disambiguation. On ToolACE-500, JTPRO updated \textbf{11\%} of tool descriptions (\textbf{55/500}) with explicit cues that better separate confusable tools. We quantify this effect using intra-group cosine similarity across \textbf{37} groups of semantically similar tools; the group with the largest improvement reduced similarity from \textbf{0.668} to \textbf{0.502} (\textbf{-16.6\%}), indicating clearer differentiation. 

%% file: sections/camera_ready/7-conclusion.tex
\section{Conclusion}
We presented \textbf{Joint Tool--Prompt Reflective Optimization (JTPRO)}, a weight-free reflective context optimization framework for improving tool-calling reliability in trace-supervised settings by jointly refining global agent instructions and per-tool schema/argument descriptions from rollout-driven feedback. JTPRO targets the two dominant failure modes in large, domain-specific tool inventories: tool mis-selection and argument mis-instantiation. The method uses reflective diagnostics to produce localized edits, maintains a candidate pool with Pareto-style selection to preserve diverse effective behaviors, and prevents context bloat by \emph{globalizing} recurring slot semantics into the instruction layer while retaining tool-specific disambiguation cues in local schemas. 
Across ToolACE tool-scaling experiments, ETID enterprise slot-filling tasks, and SEAL-Tools multi-tool calling, JTPRO improves tool selection, slot filling, and overall success relative to strong baselines including reflective prompt optimizers (e.g., GEPA), highlighting that accurate argument semantics are necessary to translate high tool selection accuracy into end-to-end tool-use success. 
Overall, these results position JTPRO as a practical reflective optimization approach for jointly adapting multiple agent operating components, global instructions and tool schemas, to improve reliable tool invocation under evolving tool inventories without model fine-tuning.

\section*{Limitations}
Our study has limitations that motivate future work.
First, our experimental scope is centered on trace-supervised tool-calling reliability. We evaluate (i) single-tool, single-slot/value cases and (ii) multi-tool \emph{parallel} calling with single-slot/value instantiation; we do not evaluate \emph{sequential} multi-tool workflows that require multi-step dependencies, intermediate state, or long-horizon planning (e.g., tool chains where earlier outputs condition later calls). Extending JTPRO to such settings will require modeling stepwise credit assignment across tool sequences and validating robustness under longer rollouts.

Second, while ETID captures complex multi-argument schemas, our current evaluation does not systematically stress \emph{deeply nested} argument structures (e.g., multi-layer JSON objects, lists-of-objects with constraints, or schema-dependent composition rules) at scale; future benchmarks should include nested-slot correctness and structure-aware metrics beyond scalar slot/value matching.

Third, our evaluation focuses on call-level correctness (tool, slots, values) in trace-supervised settings rather than executing tools and verifying response-level correctness; when tool execution is available, future experiments should extend JTPRO to end-to-end evaluation that includes tool responses and downstream post-processing logic (e.g., response parsing, aggregation, and business-rule enforcement), since these components can introduce additional failure modes beyond argument correctness.

Fourth, ETID is currently not publicly released, which limits external reproducibility and benchmarking by the community.

Finally, our empirical study is limited to 3 benchmarks; future work should broaden coverage to additional public and proprietary tool-use datasets spanning more domains, tool granularity, and interaction styles, to better characterize generalization under diverse tool inventories and schema conventions.

%% file: sections/camera_ready/9_appendix.tex
\section{Method prompts and details}

\subsection{Update Global Instructions and Tool Revisions Prompt}
\begin{tcolorbox}[
  breakable,
  enhanced,
  colback=white,
  colframe=black!60,
  boxrule=0.6pt,
  arc=2pt,
  left=8pt,right=8pt,top=8pt,bottom=8pt,
  fonttitle=\bfseries,
  title={Prompt Template: Propose Updated Global Instructions and Tool Revisions (JTPRO Reflector)},
]
\small
\textit{Goal: produce clean, minimal updates to the global instructions and only the tools that require revision, based on the feedback trace.}

\vspace{6pt}
\textbf{Task.} Update the global instructions and tool descriptions using the feedback on the current context.

\vspace{4pt}
\textbf{Current Global Instructions}\\
\texttt{<curr\_instructions>}

\vspace{4pt}
\textbf{Current Tool Definitions (Full List)}\\
\texttt{<tools\_list\_to\_update>}

\vspace{6pt}
\textbf{Objective.} Produce:
\begin{itemize}\itemsep2pt
    \item \texttt{global\_instructions}: updated system-level guidance.
    \item \texttt{example\_specific\_instructions}: batch-specific guidance to append to prior examples/hints.
    \item \texttt{tool\_revisions}: \underline{only} the tools you modified (not the entire tool list).
\end{itemize}

\vspace{4pt}
\textbf{Feedback Trace}\\
\texttt{<dataset\_with\_feedback>}\\
For each example, the feedback indicates whether the model:
\begin{itemize}\itemsep2pt
    \item failed to call a tool that should have been called, or
    \item called the wrong tool instead of the correct one, or
    \item selected the correct tool but produced incorrect \texttt{action\_inputs} (missing/wrong parameters), or
    \item selected the correct parameters but assigned incorrect slot names/values (formatting/value errors).
\end{itemize}

\vspace{6pt}
\textbf{Instructions Revision Rules (Important)}
\begin{itemize}\itemsep2pt
    \item The instructions may already contain prior revisions and examples.
    \item Always preserve previously incorporated \texttt{example\_specific\_instructions} and example hints; do not alter them.
    \item Modify existing \texttt{example\_specific\_instructions} only if there is a direct conflict with the current feedback.
    \item Add new guidance as bullet points appended to the existing list under \texttt{example\_specific\_instructions}.
\end{itemize}

\vspace{6pt}
\textbf{Tool Revision Rules (Important)}
\begin{itemize}\itemsep2pt
    \item If an answer is marked wrong, check two cases:
    \begin{itemize}\itemsep2pt
        \item \textbf{Case 1 (Model error):} If a tool-selection error occurred (a tool should have been chosen but was not, or was chosen but should not have been), you \emph{must} revise the relevant tool description(s) to make correct usage clearer. Otherwise, you may leave the tool unchanged.
        \item \textbf{Case 2 (Documentation/data issue):} Tool documentation may be ambiguous or incomplete, and ground-truth traces may contain incorrect tool arguments or values. If you detect such issues, revise the tool documentation to remove ambiguity so future runs avoid the same failure.
    \end{itemize}
    \item Return \underline{only} the tools you modified from the provided tool list.
\end{itemize}

\vspace{6pt}
\textbf{Output Format (Strict)}
\begin{itemize}\itemsep2pt
    \item Return a single JSON object immediately after the literal text \texttt{Answer:}
    \item Do not add any extra text before or after the JSON.
\end{itemize}

\vspace{4pt}
\textbf{Required JSON Schema}
\begin{quote}\small\ttfamily
\{\\
\ \ "global\_instructions": "UPDATED GLOBAL INSTRUCTIONS HERE",\\
\ \ "example\_specific\_instructions": "UPDATED INSTRUCTIONS AS PER THE CURRENT BATCH",\\
\ \ "tool\_revisions": [\\
\ \ \ \ \{\\
\ \ \ \ \ \ "name": "<tool1\_name>",\\
\ \ \ \ \ \ "description": "<tool1\_description>",\\
\ \ \ \ \ \ "parameters": \{\\
\ \ \ \ \ \ \ \ "type": "dict",\\
\ \ \ \ \ \ \ \ "properties": \{\\
\ \ \ \ \ \ \ \ \ \ "<property1>": \{"description": "<updated tool1\_property1\_description>", "type": "<property1\_type; same as original>"\},\\
\ \ \ \ \ \ \ \ \ \ "<property2>": \{"description": "<updated tool1\_property2\_description>", "type": "<property2\_type; same as original>"\}\\
\ \ \ \ \ \ \ \ \},\\
\ \ \ \ \ \ \ \ "required": ["<property1; required parameters; identical to original>"]\\
\ \ \ \ \ \ \},\\
\ \ \ \ \ \ "required": null\\
\ \ \ \ \}\\
\ \ ]\\
\}\\
\end{quote}

\textbf{Final constraint:} \texttt{Answer:} must be followed by \emph{only} the JSON object.
\end{tcolorbox}

\captionof{figure}{\textbf{JTPRO reflector prompt for proposing new global instructions and tool candidates.} The reflector updates system-level guidance and selectively revises only the implicated tools/slots based on rollout feedback, while preserving prior batch-specific examples and enforcing a strict JSON-only output format.}
\label{fig:prompt_tool_candidate_proposal}


\begin{tcolorbox}[
  breakable,
  enhanced,
  colback=white,
  colframe=black!60,
  boxrule=0.6pt,
  arc=2pt,
  left=8pt,right=8pt,top=8pt,bottom=8pt,
  fonttitle=\bfseries,
  title={Prompt Template: Merge Draft Instructions with the Global Best (MergeWithBest)},
]
\small
\textbf{Role.} You are the \emph{instruction merger} in agent. Your job is to combine a draft update with the current best global instructions into a single improved instruction prompt.

\vspace{6pt}
\textbf{Inputs}
\begin{itemize}\itemsep2pt
    \item \textbf{Global best instructions} ($P^\star$): \texttt{<best\_global\_instructions>}
    \item \textbf{Draft instructions from current rollout} ($P^d$): \texttt{<draft\_global\_instructions>}
    \item \textbf{(Optional) Newly added tools since $P^\star$}: \texttt{<new\_tools\_summary>} \\
    (Names + 1--2 lines per tool describing the new capability.)
\end{itemize}

\vspace{4pt}
\textbf{Objective.} Produce merged instructions $P'$ that:
\begin{itemize}\itemsep2pt
    \item preserve stable, broadly useful guidance from $P^\star$ (the ``growing playbook''),
    \item incorporate \emph{new} and \emph{validated} guidance from $P^d$ \emph{additively},
    \item remain concise and non-redundant (avoid restating the same rule twice),
    \item support incremental toolset growth: keep cross-cutting rules stable while adding any new decision rules introduced by newly appended tools.
\end{itemize}

\vspace{6pt}
\textbf{Merge Rules (Strict)}
\begin{itemize}\itemsep2pt
    \item \textbf{Do not overwrite}: never delete a rule from $P^\star$ unless $P^d$ provides a clearly conflicting correction.
    \item \textbf{Prefer generalization}: if $P^d$ adds a rule that generalizes an existing one, keep the generalized version and remove the narrower duplicate.
    \item \textbf{Resolve conflicts explicitly}: if a draft rule contradicts an existing rule, keep the version that is more precise and operational (clear triggers, clear expected behavior), and remove the other.
    \item \textbf{Tool-growth compatibility}: if a draft rule is specific to a newly added tool, include it only if it can be stated as a general decision rule (when-to-use / how-to-fill), otherwise keep it minimal and non-invasive.
    \item \textbf{No tool merging}: do not rename, alias, or merge tools; only adjust global instruction text.
\end{itemize}

\vspace{6pt}
\textbf{Output Format (Strict)}
\begin{itemize}\itemsep2pt
    \item Return \emph{only} the merged global instructions $P'$ as plain text.
    \item No JSON. No commentary. No additional sections.
\end{itemize}
\end{tcolorbox}

\captionof{figure}{\textbf{MergeWithBest prompt.} The merger composes rollout-specific draft instructions $P^d$ with the current global best $P^\star$ to form $P'$, preserving stable cross-cutting guidance while integrating validated new rules. This ``growing playbook'' mechanism supports incremental toolset expansion by keeping existing conventions stable and appending new decision rules when new tools are introduced.}
\label{fig:prompt_mergebest}

\begin{tcolorbox}[
  breakable,
  enhanced,
  colback=white,
  colframe=black!60,
  boxrule=0.6pt,
  arc=2pt,
  left=8pt,right=8pt,top=8pt,bottom=8pt,
  fonttitle=\bfseries,
  title={Prompt Template: Slot-Semantics Globalization (Two-Level Context Editing)},
]

\textbf{Role.} You are a context editor for a tool-using LLM agent. You may revise (i) \emph{Global Instructions} $P$ and (ii) per-tool schemas $\{T_i\}$ (tool and argument descriptions).\\
\textbf{Objective.} Reduce repeated slot/argument guidance across tools while preserving tool-specific distinctions needed for correct tool selection and slot filling.

\vspace{6pt}
\textbf{Step 1: Scan for repeated slot semantics.}
Read each tool schema $T_i$ and its argument descriptions carefully. Identify \emph{recurring} slot conventions that appear across many tools, such as:
date/time windows, identifier formatting, numeric bounds (inclusive/exclusive), units/currency normalization, boolean/defaulting rules, pagination parameters, and sorting conventions.

\vspace{4pt}
\textbf{Step 2: Globalize shared rules.}
For each repeated convention, write a \emph{single}, canonical rule in the Global Instructions $P$ that:
(a) states the default interpretation and formatting requirements, and
(b) specifies when to apply default values versus using user-provided constraints.
The global rule should be phrased generically so it applies to any tool that contains the relevant slot(s).

\vspace{4pt}
\textbf{Step 3: Keep exceptions local.}
Do \emph{not} merge, alias, or rename tools. For each tool schema $T_i$:
\begin{itemize}\setlength{\itemsep}{1pt}
    \item Remove redundant restatements of globalized rules and replace them with a short pointer (e.g., ``See Global Instructions: \emph{[Rule Name]}'' ).
    \item If a tool requires different semantics (e.g., a different date format, special rounding, a stricter constraint), keep that information \emph{locally} in $T_i$ and explicitly mark it as an \textbf{override} of the global rule.
\end{itemize}

\vspace{4pt}
\textbf{Constraints.}
\begin{itemize}\setlength{\itemsep}{1pt}
    \item Do not change tool interfaces: do \textbf{not} add/remove arguments or invent fields.
    \item Prefer minimal, high-impact edits: globalize only clearly repetitive conventions; keep tool-unique decision rules and edge cases local.
\end{itemize}

\vspace{4pt}
\textbf{Output.}
\begin{itemize}\setlength{\itemsep}{1pt}
    \item An updated Global Instructions block to append to $P$ (named rules + concise definitions).
    \item Updated schemas for only the tools you modified (short pointers + explicit overrides).
\end{itemize}
\end{tcolorbox}

\section{Enterprise Tool-Inventory Dataset (ETID): Synthesis Methodology}
\label{app:etid_release}

\paragraph{Design objective.}
ETID can be synthesized as a privacy-preserving benchmark for tool use in realistic enterprise settings, where agents must operate over a large catalog of domain-specific tools with heterogeneous schemas, overlapping capabilities, and multi-argument interfaces. Rather than relying on proprietary traces, the dataset can be constructed from an abstracted enterprise tool inventory that preserves only the structural properties needed for evaluation: tool granularity, schema complexity, required/optional argument patterns, and regions of semantic overlap.

\paragraph{Constructing an enterprise-style tool inventory.}
A practical synthesis recipe begins by assembling tools from multiple enterprise workflows, for example, search, analytics, reporting, inventory, finance, support, scheduling, and operations. Each tool is represented by a name, a concise description, and a typed parameter schema. The inventory should be synthesized to exhibit a realistic argument distribution, with many tools requiring only a small number of fields and a meaningful long tail of more complex schemas. To preserve realistic routing failures, semantically adjacent tools should remain distinct rather than being merged or aliased; instead, they should differ in scope, triggering conditions, or argument constraints so that tool disambiguation remains a central challenge.

\paragraph{Injecting overlapping field semantics.}
To reproduce the schema redundancy typical of enterprise tool stacks, the synthesis process should explicitly reuse recurring slot families across many tools. As discussed in the global slot-semantics analysis, these families include identifier fields, date/time windows, numeric bounds, boolean flags, sorting controls, and currency/unit parameters. One effective procedure is to define a library of canonical slot families and instantiate them repeatedly across tools with minor naming variation and tool-specific overrides. For example, many tools may expose fields analogous to \texttt{startDate}, \texttt{endDate}, \texttt{rangeMinimum}, \texttt{rangeMaximum}, and associated inclusivity flags, while retaining local fields such as item names, location names, account identifiers, or business-specific selectors. This creates the heavy-tailed overlap pattern that motivates globalizing shared semantics while keeping tool-specific exceptions local.

\paragraph{Generating user requests and gold traces.}
Given the synthesized inventory, natural-language requests can be generated by sampling intents per tool and paraphrasing them across diverse linguistic forms. To make the benchmark challenging, the generator should include (i) direct requests, (ii) ambiguous requests that could plausibly match multiple nearby tools, and (iii) requests that require normalization of dates, identifiers, numeric ranges, booleans, or units. Gold traces are then constructed as structured tool calls with the correct tool and fully specified arguments, including canonical formatting and defaulting rules for recurring slot families.

\paragraph{Privacy and validation.}
Because the benchmark is synthetic, all values can be produced from controlled non-sensitive templates or vocabularies, with explicit exclusion of personally identifiable information (PII), proprietary identifiers, and real customer artifacts. A final validation stage should verify schema consistency, argument-type correctness, intended ambiguity among overlapping tools, and coverage across train, validation, and test splits. Low-shot regimes can then be created by allocating a small number of labeled examples per tool while maintaining broad tool coverage in evaluation.

\input{figures_camera_ready/tex/fig4_toolace_results_camera_ready}
\input{figures_camera_ready/tex/fig5_etid_results_camera_ready}
\section{Experimental Setup}
\label{app:exp_setup}

\subsection{Benchmark Suitability and Scope of Evaluation}
\label{app:benchmark_scope}

Our benchmark selection is guided by the core design assumption of JTPRO: prompt and schema refinements should be learned over a reusable tool set with stable schema semantics, so that improvements transfer across train, validation, and test splits rather than overfitting to individual instances. This makes benchmarks with persistent tools and large schema-rich tool inventories especially suitable for evaluating our method.

\paragraph{Why we do not use BFCL.}
BFCL defines tools on a per-instance basis rather than as a single consistent tool inventory. In practice, we observed cases where the same tool name appears with different parameter definitions across examples. This makes it difficult to construct the stable, reusable tool universe required by JTPRO without substantially modifying the dataset. For this reason, we considered BFCL unsuitable for our study setting.

\paragraph{Why $\tau$-bench is only partially aligned.}
$\tau$-bench is closer to our setting, but it uses relatively small tool inventories (e.g., 15 APIs in Retail and 13 in Airline). By contrast, our main claim concerns settings with large toolsets, where tool selection and schema-constrained slot filling become the dominant sources of failure. We therefore prioritize benchmarks that more directly stress this scaling regime.

\paragraph{Scope with respect to sequential planning.}
Our current evaluation focuses on single-tool, parallel multi-tool, and schema-constrained argument instantiation settings. It does not yet evaluate long-horizon sequential workflows in which the output of one tool determines the choice or arguments of a subsequent tool. We view this as an important next step, but also as a distinct source of difficulty from the large-inventory tool selection and slot-filling problems targeted by JTPRO. Extending the framework to sequential planning benchmarks is a natural direction for future work.

\subsection{Evaluation protocol and reporting.}
To reduce variance from stochastic decoding and optimization dynamics, all results are aggregated over multiple independent runs. Unless otherwise stated, each experiment is repeated \textbf{5--10 times} and we report the \textbf{average} performance across runs. For fair comparison, we run \textbf{JTPRO and GEPA under matched optimization budgets}: both methods use the same maximum number of rollouts and identical optimization settings (including the same minibatch size and the same reflector configuration). Across all experiments, we use \textbf{o3-mini} as the reflector model and set the \textbf{LLM temperature parameter to 1}.

\input{figures_camera_ready/tex/fig_7.1_global_semantics_plot}
\input{figures_camera_ready/tex/fig_7.2_global_semantics_examples}

\input{figures_camera_ready/tex/fig7_slot_improvement_percentage}
\input{figures/fig8_slot_imprv_example_toolace}
\input{figures/fig9_slot_imprv_example_etid}
\input{figures_camera_ready/tex/fig6_jtpro_convergence_camera_ready}

\section{Additional Figure Discussion}
\label{app:fig_discussion}

\subsection{Repetitive slot semantics across tools}
Figures~\ref{fig:global_semantics_plot} and~\ref{fig:global_semantics_example} motivate \textsc{GlobalizeSlots}. Figure~\ref{fig:global_semantics_plot} shows a heavy-tailed distribution of parameter families across the tool inventory: a small number of recurring semantic classes, especially identifier and date/time fields, appear across a large fraction of tools (up to 77/124), while numeric bounds, boolean flags, sorting parameters, and related fields also recur repeatedly. This pattern indicates that many tools restate nearly identical guidance for formatting and interpretation, such as ISO-8601 date handling, inclusive versus exclusive numeric bounds, defaulting behavior, boolean semantics, and currency normalization.

Figure~\ref{fig:global_semantics_example} shows how JTPRO converts this redundancy into a compact global rule layer. Instead of repeating the same instructions inside each tool schema, JTPRO lifts shared conventions into named global rules such as \emph{DateTime Fields}, \emph{Numeric Bounds}, \emph{Boolean Parameters}, \emph{Sorting Conventions}, and \emph{Currency and Units}, and replaces duplicated local text with short pointers to those rules. In the illustrated \texttt{ItemOverageRequest} schema, for example, \texttt{startDate} and \texttt{endDate} now refer to the global date/time rule, \texttt{rangeMinimum} and \texttt{rangeMaximum} refer to the global numeric-bounds rule, and \texttt{rangeMinimumInclusive} and \texttt{rangeMaximumInclusive} refer to the global boolean rule, while tool-specific fields such as \texttt{locationName} and \texttt{itemName} remain local. This two-level organization improves slot filling by enforcing consistent semantics across tools, reducing duplicated and potentially conflicting wording, and preserving schema space for the tool-specific exceptions and disambiguation cues that matter for correct invocation.

\subsection{Figure~\ref{fig:toolace_scaling_bars}: ToolACE scaling results}
Figure~\ref{fig:toolace_scaling_bars} evaluates robustness under tool-universe growth (500 vs.\ 1000 tools). The dominant failure mode under scaling is reduced \textsc{TSA}: as inventories expand, overlapping tool descriptions and increased distractors cause more routing errors, which then cascade into lower \textsc{OSR}. GEPA partially mitigates this via global instruction refinement, but it does not directly repair tool-local ambiguity or slot semantics. JTPRO delivers the most consistent \textsc{OSR} gains because it \textbf{jointly} revises global policies \emph{and} the specific tool/slot descriptions implicated by observed failures, improving both selection and downstream argument correctness.

\subsection{Figure~\ref{fig:etid_bars}: ETID performance across supervision levels}
Figure~\ref{fig:etid_bars} studies data efficiency on ETID under Train-1ex/2ex/4ex regimes. Across models, baselines often achieve relatively strong \textsc{TSA} but substantially lower \textsc{OSR}, indicating that \textbf{slot/value instantiation} is the primary bottleneck under complex schemas. JTPRO improves \textsc{SFA} (conditional on \textsc{TSA}) and therefore consistently lifts \textsc{OSR} across supervision levels, reflecting that many ETID failures stem from underspecified or inconsistent argument semantics that can be corrected through targeted tool/slot documentation edits plus strengthened global tool-calling rules.

\subsection{Figure~\ref{fig:slot_gain_rate}: Per-example slot-filling improvement rate}
Figure~\ref{fig:slot_gain_rate} reports the fraction of test instances for which JTPRO improves per-query slot/value correctness over the baseline. The gains are larger on ETID (complex schemas) than on ToolACE, aligning with the hypothesis that real-world OSR is often bottlenecked by argument instantiation even after correct tool selection. Importantly, the improvements occur on a non-trivial fraction of held-out examples across all evaluated models, suggesting that joint context refinement yields robust, example-level corrections rather than isolated wins.

\subsection{Figure~\ref{fig:toolace500_gpt5_example_slot_impr}: ToolACE-500 example-wise corrections (GPT-5)}
Figure~\ref{fig:toolace500_gpt5_example_slot_impr} provides an example-wise view of slot/value corrections on ToolACE-500 for GPT-5. Each bar corresponds to a test instance, contrasting baseline vs.\ JTPRO slot correctness. Overall, JTPRO improves slot correctness on \textbf{26/121} examples (\textbf{21.48\%}). This plot highlights that improvements are distributed across the test set (rather than concentrated in a single cluster), consistent with JTPRO correcting recurring slot conventions and tool-specific documentation ambiguities that manifest in diverse queries.
\subsection{Figure~\ref{fig:etid_gpt5_example_slot_impr}: ETID example-wise corrections (GPT-5)}
Figure~\ref{fig:etid_gpt5_example_slot_impr} shows the analogous example-wise comparison for ETID (GPT-5). JTPRO improves slot correctness on \textbf{94/403} examples (\textbf{23.33\%}), reinforcing that complex multi-argument schemas benefit strongly from (i) tightening global tool-calling policies (e.g., required-field completeness, no hallucinated keys) and (ii) clarifying per-tool parameter semantics. Together with Figures~\ref{fig:etid_bars} and \ref{fig:slot_gain_rate}, this example-level view supports the claim that \textsc{SFA} is a major driver of end-to-end \textsc{OSR} gains on ETID.

\subsection{Figure~\ref{fig:jtpro_convergence}: Convergence behavior}
Figure~\ref{fig:jtpro_convergence} plots validation \textsc{OSR} over JTPRO iterations for three base models, with $\star$ denoting the final test \textsc{OSR} obtained using the best validation-selected context. The curves show rapid early gains followed by saturation, consistent with the reflector first correcting high-impact, systematic errors (e.g., frequent tool confusions, missing required slots, formatting/defaulting mistakes) and later iterations focusing on smaller refinements. The separation between validation trajectories and the final test markers indicates that improvements transfer to held-out queries rather than merely optimizing minibatch idiosyncrasies.

\subsubsection{Embedding-Based Disambiguation Metric}

\subsection{Tool Description Disambiguation Analysis}
\label{app:disambiguation}

We study how joint optimization improves tool selection by analyzing semantic changes in tool descriptions on the ToolAce-500 benchmark.

\subsubsection{Description Enrichment}

The optimizer modifies 55 out of 500 tool descriptions (11\%), increasing the average description length from 86.1 to 100.1 characters (+16.3\%). These edits primarily target \emph{disambiguation}, explicitly differentiating tools with overlapping semantics.

\paragraph{Example: \texttt{search} vs.\ \texttt{web\_search}}
\vspace{2pt}
\begin{itemize}[leftmargin=1.2em, noitemsep]
    \item \textbf{search (before):} ``Perform Google search and get results.''
    \item \textbf{search (after):} ``Perform Google search with advanced locale controls (gl/hl), country restrictions (cr), and time filters (tbs). \emph{NOT for general web article or paper discovery, prefer \texttt{web\_search} for generic queries.}''
    \item \textbf{web\_search (after):} ``Search the web for relevant pages. \emph{PREFERRED for general-purpose web, article, and paper discovery. Do not confuse with similarly named \texttt{search} tools.}''
\end{itemize}

\subsubsection{Confusable Tool Groups}

We identify tools sharing common name prefixes (e.g., \texttt{get\_user\_*}, \texttt{get\_all\_*}) as potentially confusable. This yields 37 groups comprising 109 tools, representing high-risk ambiguity regions where models frequently select semantically similar but incorrect tools.

To quantify disambiguation quality, we compute pairwise cosine similarity between tool descriptions within each confusable group using sentence embeddings (\texttt{all-MiniLM-L6-v2}). Lower intra-group similarity indicates stronger semantic separation.

\begin{table}[t]
    \centering
    \small
    \setlength{\tabcolsep}{4pt}
    \begin{tabular}{lccc}
        \toprule
        \textbf{Tool Group} & \textbf{Before} & \textbf{After} & \textbf{$\Delta$} \\
        \midrule
        \texttt{get\_ip\_*} (2) & 0.668 & 0.502 & $-$0.166 \\
        \texttt{get\_page\_*} (2) & 0.849 & 0.736 & $-$0.113 \\
        \texttt{get\_languages\_*} (2) & 0.676 & 0.584 & $-$0.092 \\
        \texttt{get\_trending\_*} (3) & 0.437 & 0.377 & $-$0.060 \\
        \texttt{search\_by\_*} (2) & 0.281 & 0.265 & $-$0.016 \\
        \bottomrule
    \end{tabular}
    \caption{Intra-group cosine similarity (lower indicates better disambiguation) for the most improved confusable tool groups. Parentheses denote group size.}
    \label{tab:disambiguation}
\end{table}

The \texttt{get\_ip\_*} group exhibits the largest improvement, with similarity reduced from 0.668 to 0.502. This change reflects the addition of explicit preference guidance, e.g.,
``\emph{Preferred for general IP geolocation requests. Use this instead of \texttt{get\_geolocation\_by\_ip} unless extended fields are required.}''

\subsubsection{Disambiguation Patterns Learned}

Across the 55 modified tools, we observe four recurring disambiguation strategies:

\begin{enumerate}[leftmargin=1.2em, noitemsep]
    \item \textbf{Parameter format guidance} (28 tools): e.g., ``Pass \texttt{country\_code} as a 2-letter lowercase ISO code.''
    \item \textbf{Explicit preference signals} (9 tools): e.g., ``PREFERRED for\ldots''
    \item \textbf{Negative constraints} (5 tools): e.g., ``NOT for general web article discovery.''
    \item \textbf{Cross-tool references} (3 tools): e.g., ``Use this instead of \texttt{get\_geolocation\_by\_ip}.''
\end{enumerate}

\subsubsection{Summary Statistics}

\begin{table}[t]
    \centering
    \small
    \setlength{\tabcolsep}{5pt}
    \begin{tabular}{lr}
        \toprule
        \textbf{Metric} & \textbf{Value} \\
        \midrule
        Total tools analyzed & 500 \\
        Modified descriptions & 55 (11.0\%) \\
        Avg.\ length (before) & 86.1 chars \\
        Avg.\ length (after) & 100.1 chars \\
        Relative increase & +16.3\% \\
        \midrule
        Confusable groups & 37 \\
        Confusable tools & 109 \\
        Groups improved & 6 (16.2\%) \\
        Avg.\ similarity reduction & 0.012 \\
        \bottomrule
    \end{tabular}
    \caption{Summary of tool description disambiguation on ToolAce-500.}
    \label{tab:disambig_summary}
\end{table}

Overall, joint optimization learns to resolve tool ambiguity through targeted description edits, complementing instruction-level optimization and improving tool selection robustness.

\begin{figure*}[t]
    \centering
    \includegraphics[width=0.81\textwidth]{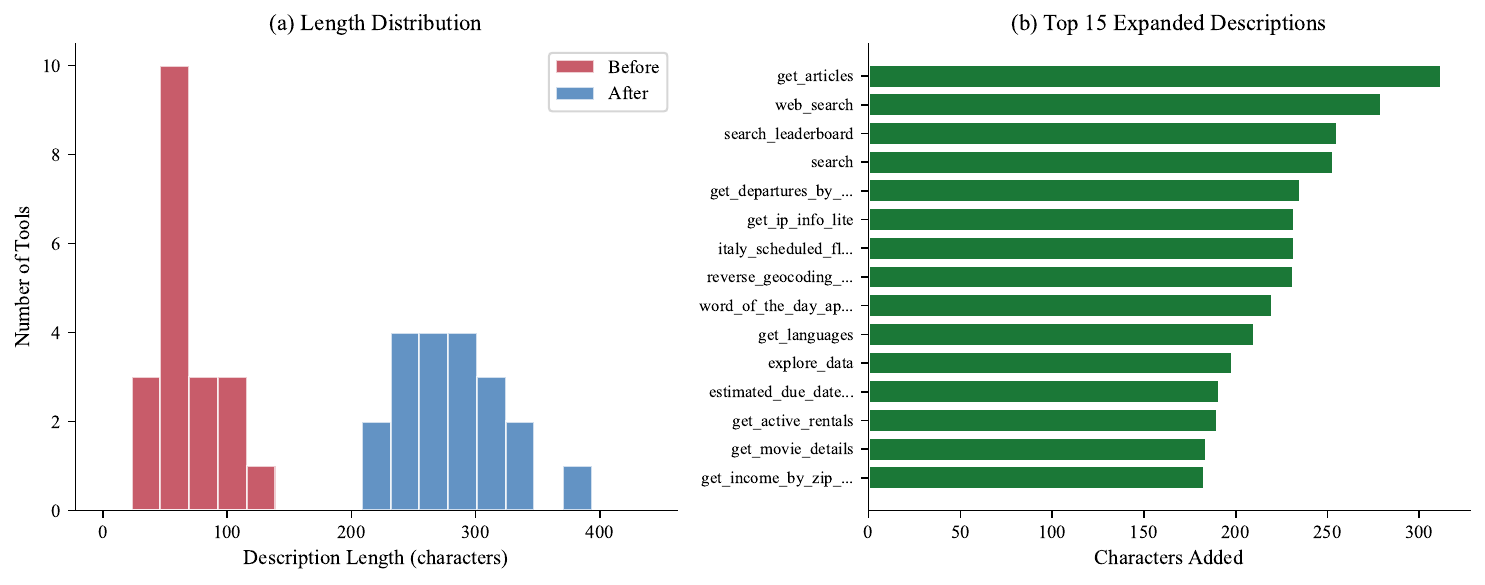}
    \caption{
    Tool description length analysis.
    (a) Distribution of description lengths before and after optimization.
    (b) Per-tool length changes for the 55 modified tools.
    }
    \label{fig:desc_length}
\end{figure*}


\begin{figure*}[t]
    \centering
    \includegraphics[width=\linewidth]{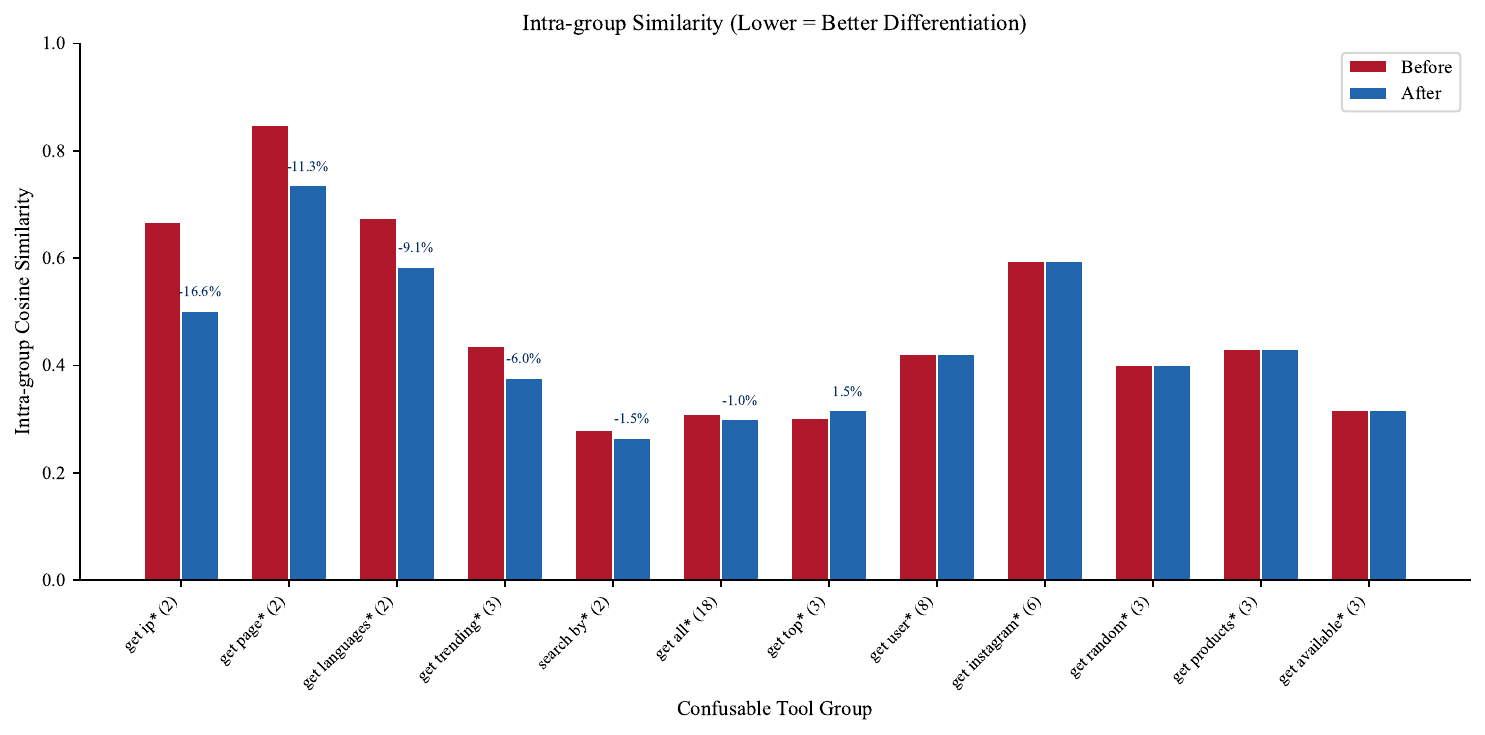}
    \caption{
    Intra-group cosine similarity for the top 15 confusable tool groups.
    Lower values indicate stronger semantic differentiation.
    }
    \label{fig:sim_improvement}
\end{figure*}


%% file: figures_camera_ready/tex/fig4_toolace_results_camera_ready.tex
\begin{figure*}[t]
\centering
\includegraphics[width=\textwidth]{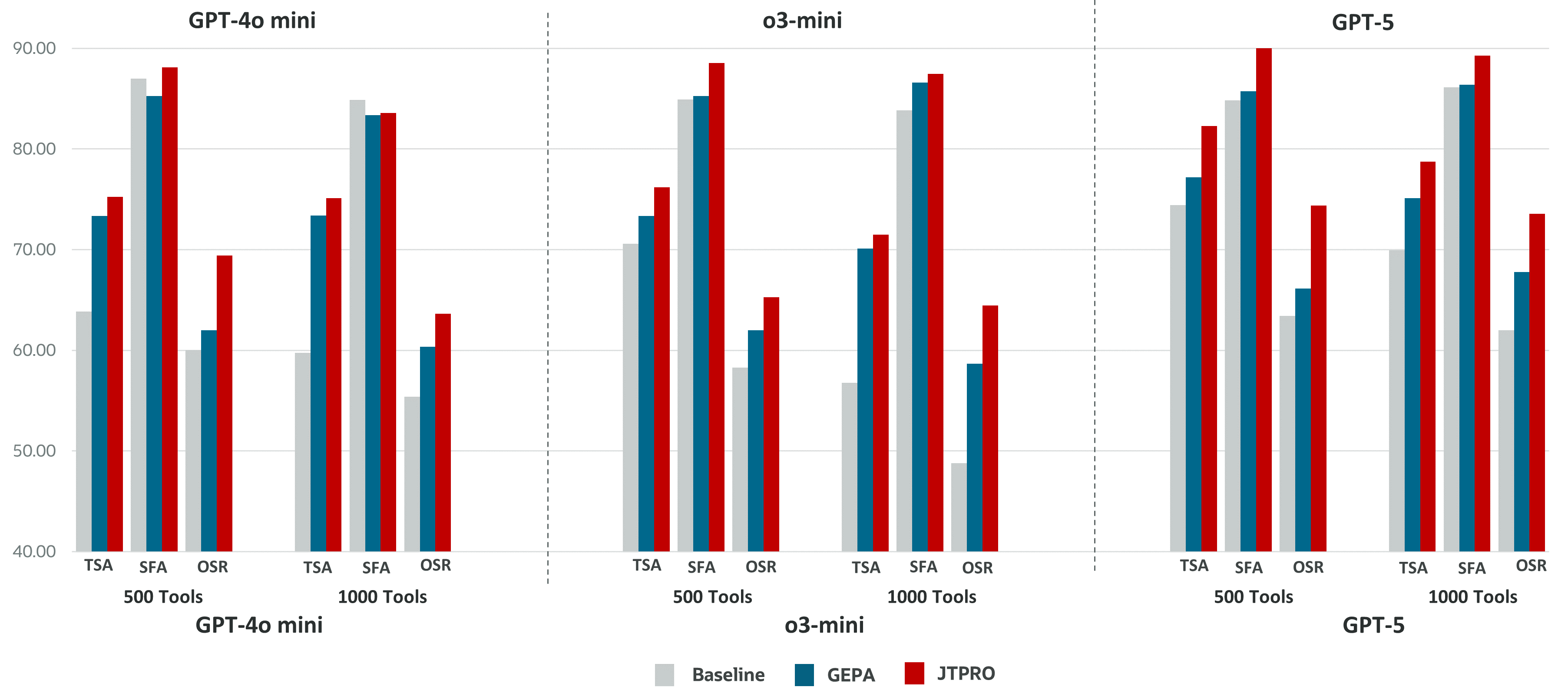}
\caption{\textbf{ToolACE scaling results across models and metrics.} For each model, we report \textsc{TSA}, \textsc{SFA} (conditional on correct tool), and \textsc{OSR} at 500 and 1000 tools. Tool-universe growth primarily reduces \textsc{TSA} for the baseline, which cascades to lower \textsc{OSR}; GEPA partially mitigates this via global instruction refinement, while JTPRO provides the most consistent improvements in \textsc{OSR} by jointly refining global instructions and tool/slot descriptions.}
\label{fig:toolace_scaling_bars}
\end{figure*}

%% file: figures_camera_ready/tex/fig5_etid_results_camera_ready.tex
\begin{figure*}[t]
\centering
\includegraphics[width=\textwidth]{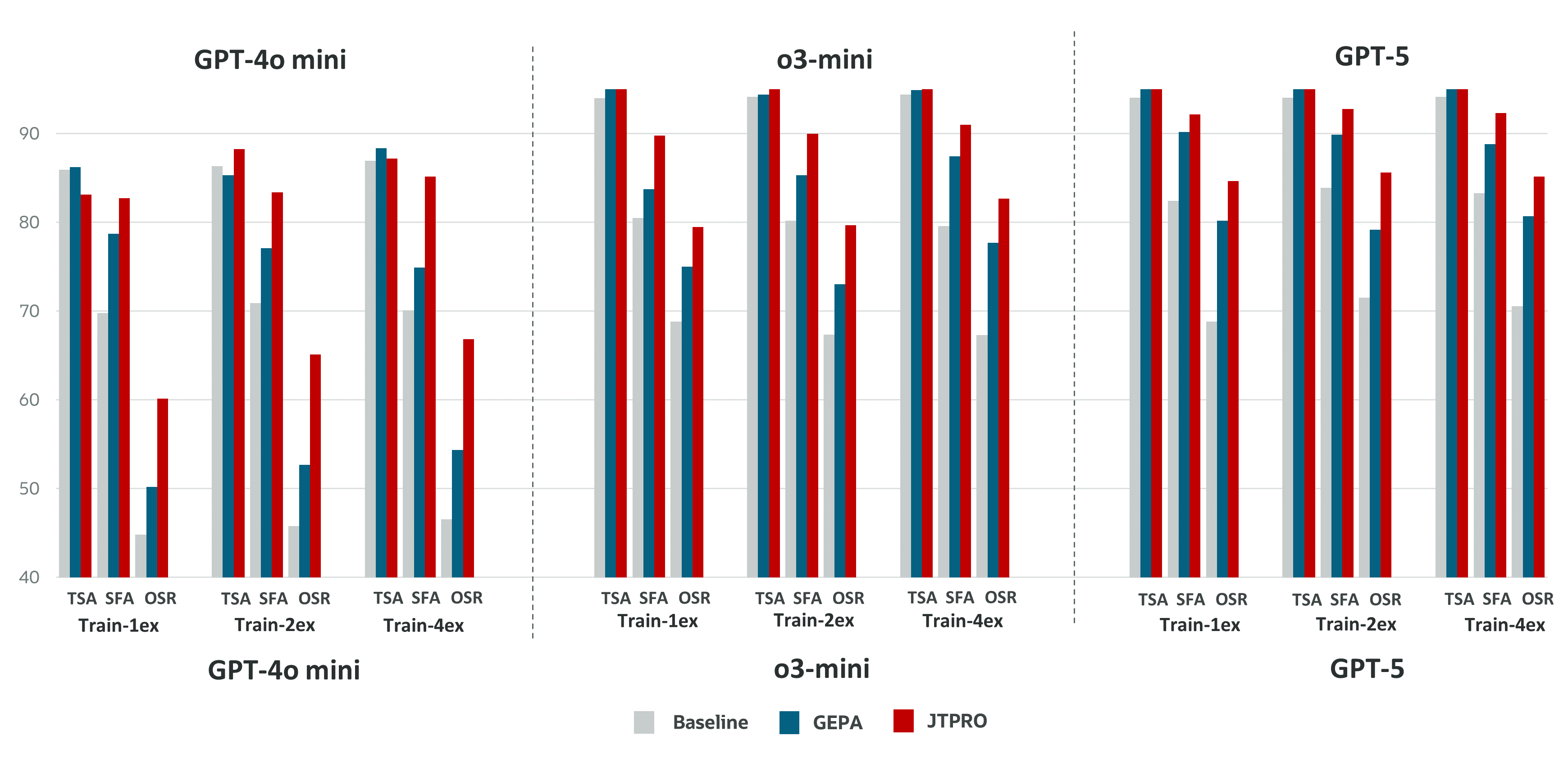}
\caption{\textbf{ETID performance across supervision levels.} Grouped bars show \textsc{TSA}, \textsc{SFA} (conditional on \textsc{TSA}), and \textsc{OSR} for three models under Train-1ex/2ex/4ex regimes. Baselines achieve high \textsc{TSA} but substantially lower \textsc{OSR}, revealing slot/value errors as the dominant failure mode; JTPRO improves \textsc{SFA} and therefore \textsc{OSR} consistently across regimes, while GEPA primarily improves \textsc{TSA} for larger models.}
\label{fig:etid_bars}
\end{figure*}

%% file: figures_camera_ready/tex/fig_7.1_global_semantics_plot.tex
\begin{figure*}[htbp] 
    \centering
    \includegraphics[width=0.7\textwidth]{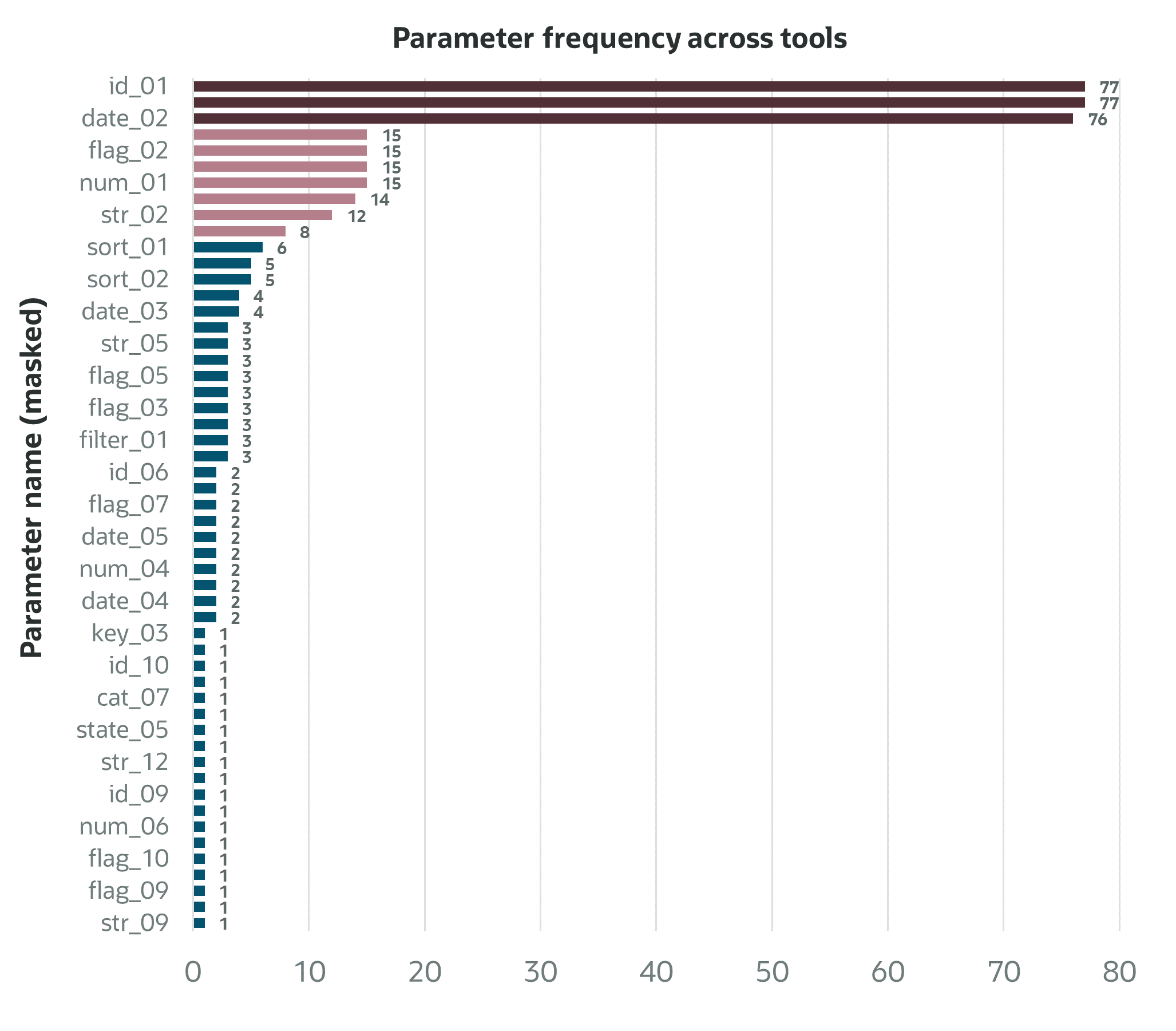} 
    \caption{Parameter frequency across tools. The distribution is heavy-tailed: a small number of parameter families, especially identifier and date/time fields, recur across a large fraction of the tool inventory (up to 77/124 tools), while numeric, boolean, sorting, and related fields also appear repeatedly. This motivates lifting shared slot semantics into a global instruction layer rather than restating them in each tool schema.}
\label{fig:global_semantics_plot}

\end{figure*}

%% file: figures_camera_ready/tex/fig_7.2_global_semantics_examples.tex
\begin{figure*}[htbp] 
    \centering
    \includegraphics[width=0.9\textwidth]{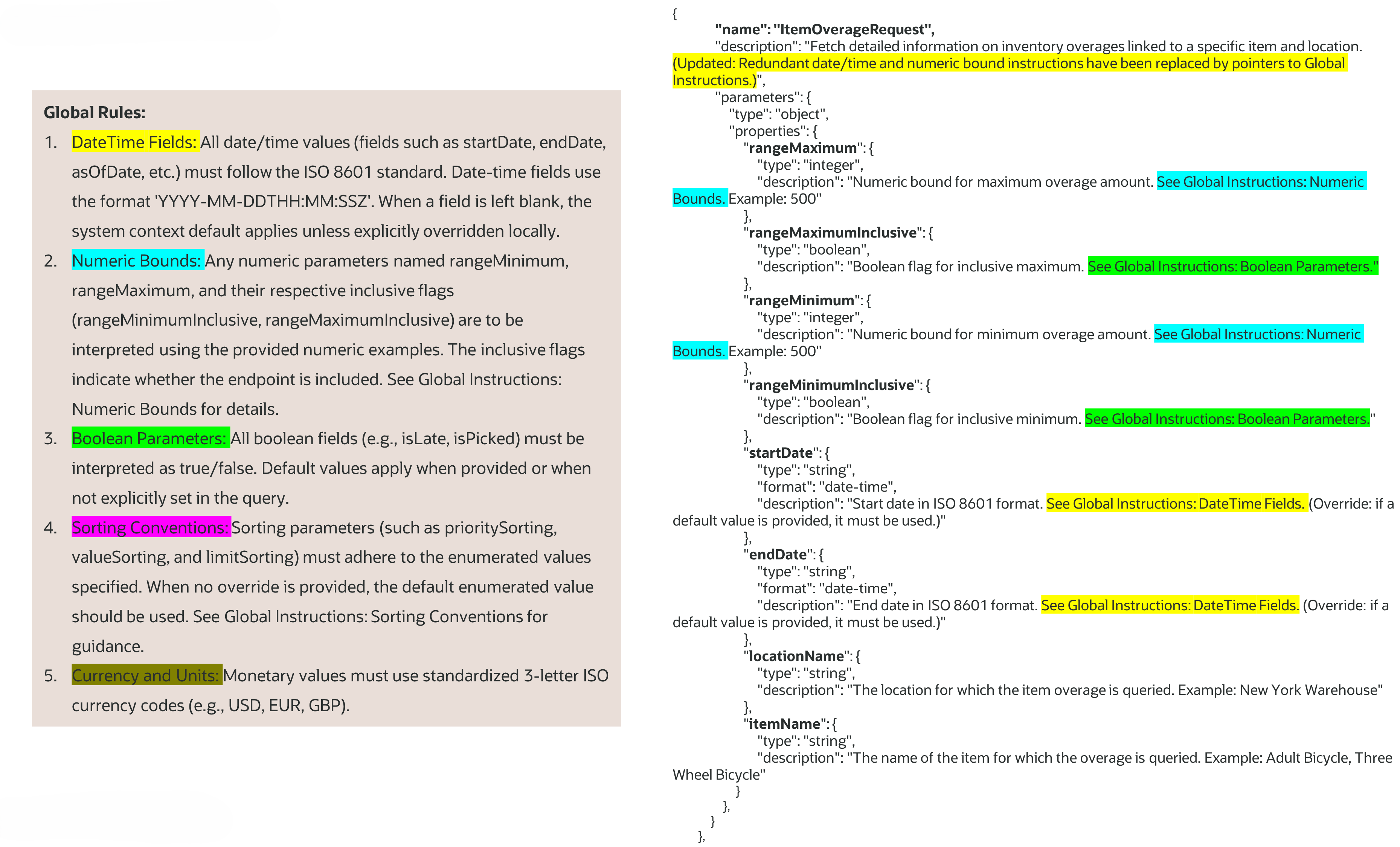} 
    \caption{Example of globalizing slot semantics. JTPRO moves repeated guidance for date/time formatting, numeric bounds, boolean interpretation, sorting conventions, and currency normalization into named global rules, and replaces redundant tool-local descriptions with short pointers to those rules. In the example \texttt{ItemOverageRequest} schema, shared fields such as \texttt{startDate}, \texttt{endDate}, \texttt{rangeMinimum}, \texttt{rangeMaximum}, and their inclusive flags reference global instructions, while tool-specific fields and local overrides remain in the tool schema.}
\label{fig:global_semantics_example}

\end{figure*}

%% file: figures_camera_ready/tex/fig7_slot_improvement_percentage.tex
\begin{figure*}[htbp] 
    \centering
    \includegraphics[width=0.81\textwidth]{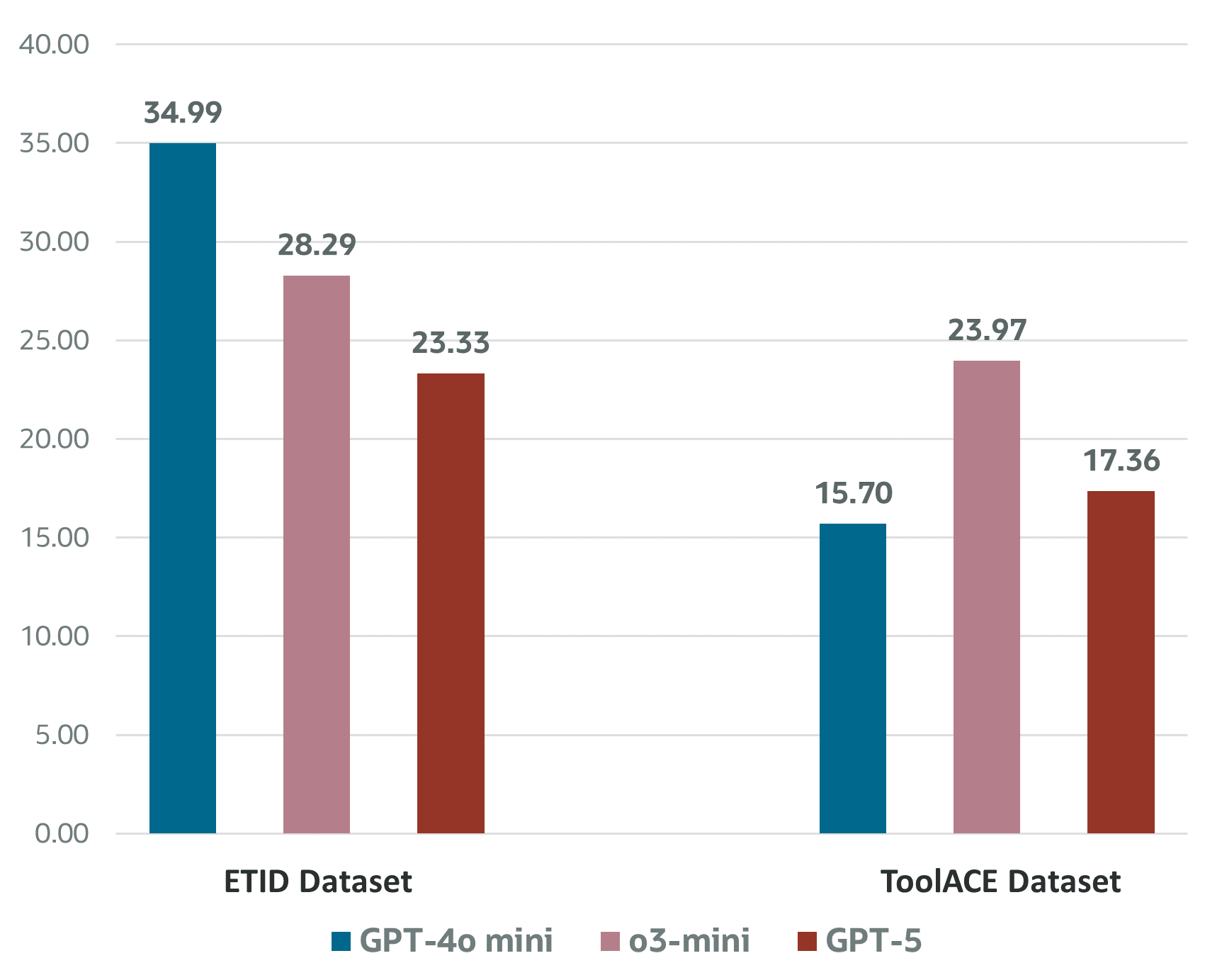} 
    \caption{\textbf{Per-example slot-filling improvements from JTPRO.}
For each base model and dataset, we report the \emph{average percentage of test instances} on which slot filling is more accurate after JTPRO optimization than the corresponding baseline (i.e., per-query slot/value correctness improves). Gains are larger on the complex slot-filling \textsc{ETID} benchmark (e.g., 34.99\% for GPT-4o mini) and remain substantial on ToolACE (e.g., 23.97\% for o3-mini), indicating that JTPRO improves argument instantiation on a non-trivial fraction of held-out queries across models.}
\label{fig:slot_gain_rate}

\end{figure*}

%% file: figures/fig8_slot_imprv_example_toolace.tex
\begin{figure*}[htbp] 
    \centering
    \includegraphics[width=0.95\textwidth]{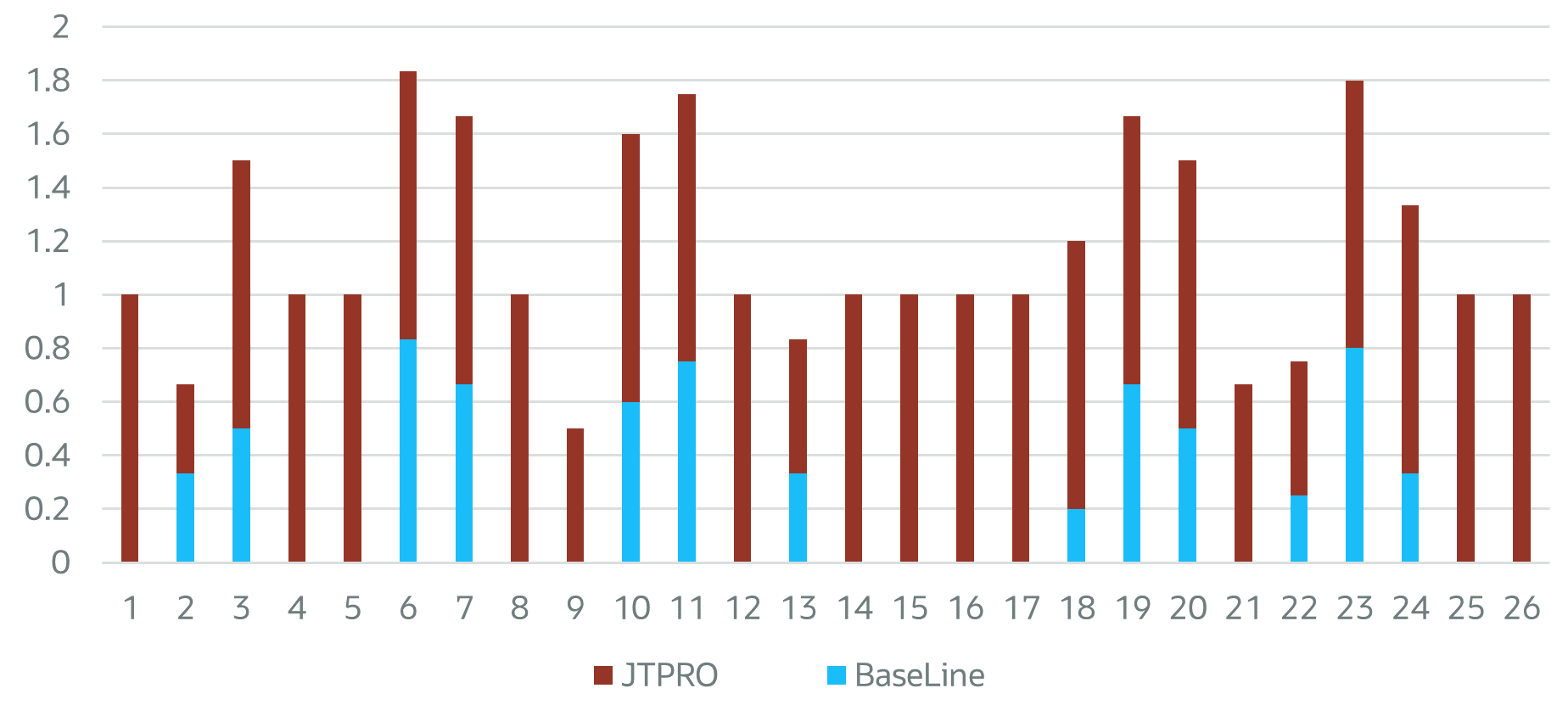} 
    \caption{\textbf{Per-example slot/value corrections after JTPRO (ToolACE-500, GPT-5).}
Example-wise comparison of slot-filling outcomes on the ToolACE test set with 500 tools for GPT-5: each bar corresponds to a test instance (x-axis indices), highlighting instances where JTPRO fixes previously incorrect slot/value instantiations relative to the baseline. Overall, JTPRO improves slot correctness on \textbf{26 out of 121} test examples (\textbf{21.48\%}).}
\label{fig:toolace500_gpt5_example_slot_impr}

\end{figure*}

%% file: figures/fig9_slot_imprv_example_etid.tex
\begin{figure*}[htbp] 
    \centering
    \includegraphics[width=0.9\textwidth]{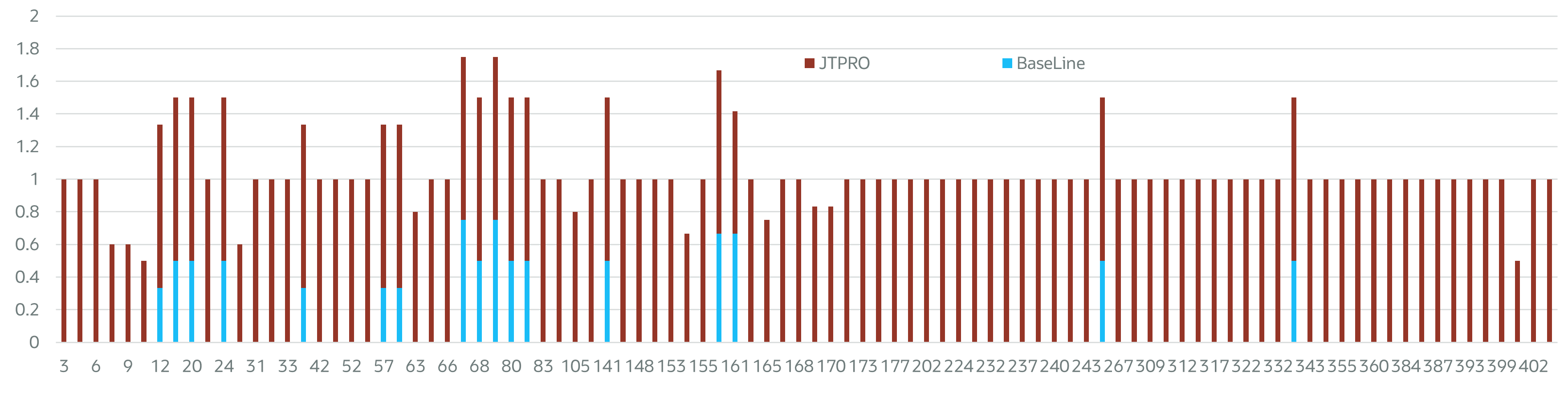} 
    \caption{\textbf{Per-example slot/value corrections after JTPRO (ETID, GPT-5).}
Example-wise comparison of slot-filling outcomes on the ETID test set for GPT-5: each bar corresponds to a test instance (x-axis indices), highlighting instances where JTPRO fixes previously incorrect slot/value instantiations relative to the baseline. Overall, JTPRO improves slot correctness on \textbf{94 out of 403} test examples (\textbf{23.33\%}).}
\label{fig:etid_gpt5_example_slot_impr}
\end{figure*}

%% file: figures_camera_ready/tex/fig6_jtpro_convergence_camera_ready.tex
\begin{figure*}[htbp] 
    \centering
    \includegraphics[width=0.90\textwidth]{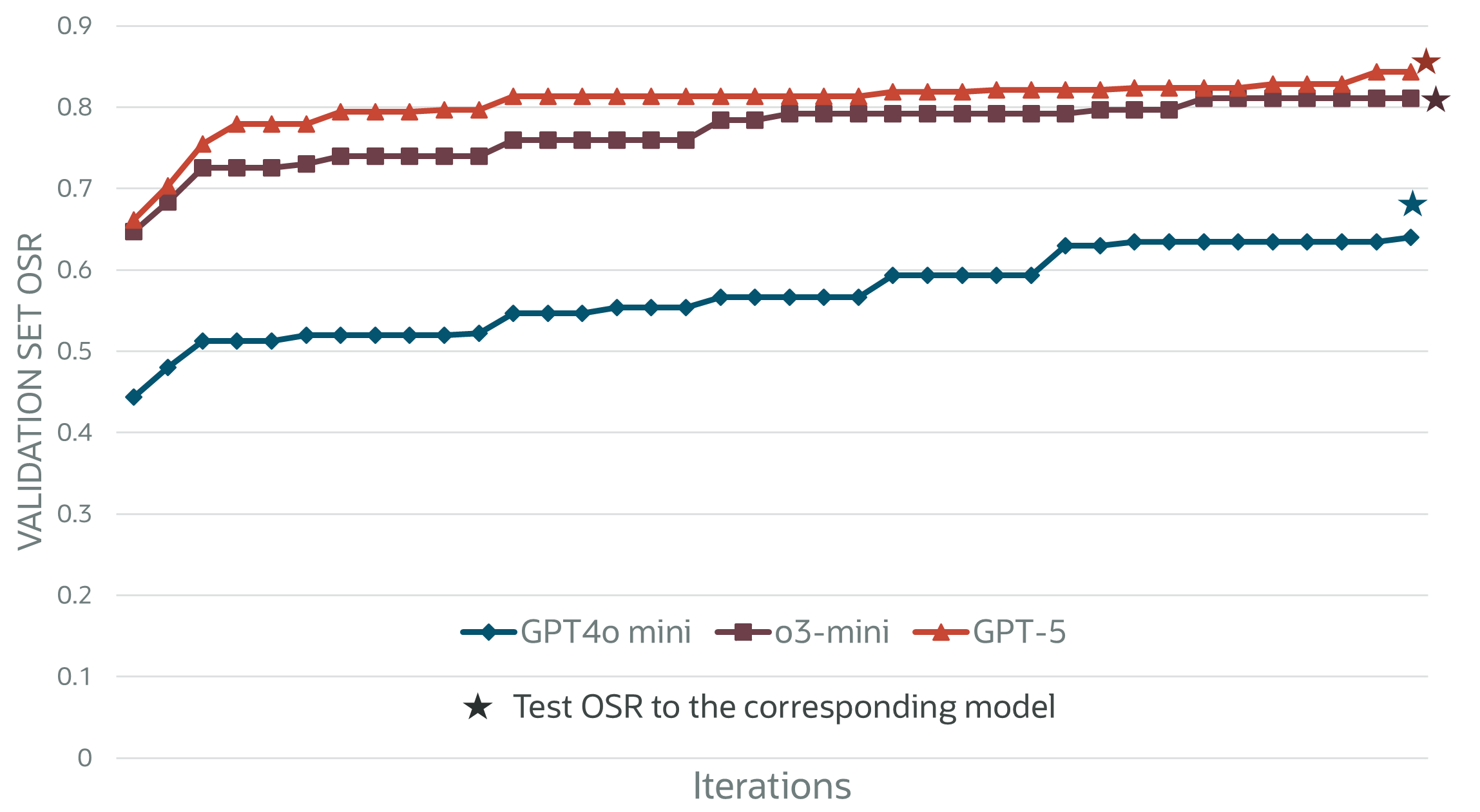} 
    \caption{\textbf{JTPRO convergence on validation OSR.} Validation OSR over optimization iterations for three base models (GPT-4o mini, o3-mini, GPT-5); $\star$ marks final test OSR for the best validation-selected context. OSR rises quickly in early iterations and then plateaus, indicating rapid correction of high-impact errors followed by smaller refinements that transfer to held-out data.}
    \label{fig:jtpro_convergence}
\end{figure*}